\pdfoutput=1
\documentclass{article}

\usepackage[final,main]{neurips_2026}
\PassOptionsToPackage{hidelinks,colorlinks=false,pdfborderstyle={/S/U/W 0}}{hyperref}
\usepackage[utf8]{inputenc}
\usepackage[T1]{fontenc}
\usepackage{url}
\usepackage{booktabs}
\usepackage{amsmath,amssymb,amsfonts}
\usepackage{microtype}
\usepackage{xcolor}
\usepackage{graphicx}
\usepackage{longtable}
\usepackage{multirow}
\usepackage{tikz}
\usepackage{pgfplots}
\usepackage{float}
\usepackage{needspace}
\usepackage{placeins}
\usepackage{pgfplotstable}
\usepgfplotslibrary{groupplots}
\usetikzlibrary{arrows.meta,positioning,calc,fit}
\pgfplotsset{compat=1.18}
\usepackage{hyperref}
\hypersetup{
  pdftitle={Expanding LLM Reasoning},
  pdfauthor={Rian Atri and Evan Luo}
}

\definecolor{cbblue}{RGB}{0,114,178}
\definecolor{cborange}{RGB}{230,159,0}
\definecolor{cbpurple}{RGB}{170,68,153}
\newcommand{\pp}{\,\mathrm{pp}}
\newcommand{\Eset}{\mathcal E}
\newcommand{\Top}{\operatorname{Top}}

\makeatletter
\newcommand{\SectionSpace}{\par\penalty\z@\ifdim\pagegoal<\maxdimen\ifdim\dimexpr\pagegoal-\pagetotal\relax<7\baselineskip\newpage\fi\fi}
\makeatother
\newcommand{\EqThreeGapPP}{+14.80}
\newcommand{\EqThreePositiveCells}{41}
\newcommand{\EqThreePrimaryCells}{41}
\newcommand{\MaximumObservedUtilityPP}{+21.34}
\newcommand{\PooledEligibleGini}{0.896}
\newcommand{\PooledEligibleGiniHighCells}{37}
\newcommand{\WithinExampleEligibleGini}{0.585}
\newcommand{\GiniGapCorrelation}{-0.563}
\newcommand{\RouterUniformSeeds}{25}
\newcommand{\RouterUniformWins}{25}
\newcommand{\DepthResidualA}{+0.0275}
\newcommand{\DepthResidualALow}{+0.0183}
\newcommand{\DepthResidualAHigh}{+0.0368}
\newcommand{\DepthResidualAPositiveCells}{14}
\newcommand{\DepthResidualB}{+0.0248}
\newcommand{\DepthResidualBLow}{+0.0157}
\newcommand{\DepthResidualBHigh}{+0.0340}
\newcommand{\DepthResidualBPositiveCells}{15}
\newcommand{\DepthResidualCells}{16}
\newcommand{\DepthResidualExamples}{3421}
\newcommand{\RichResidualA}{+0.03228}
\newcommand{\RichResidualAProblemLow}{+0.02409}
\newcommand{\RichResidualAProblemHigh}{+0.04048}
\newcommand{\RichResidualACellLow}{+0.01941}
\newcommand{\RichResidualACellHigh}{+0.05055}
\newcommand{\RichResidualAPositiveCells}{15}
\newcommand{\RichResidualB}{+0.02956}
\newcommand{\RichResidualBProblemLow}{+0.02111}
\newcommand{\RichResidualBProblemHigh}{+0.03818}
\newcommand{\RichResidualBCellLow}{+0.01730}
\newcommand{\RichResidualBCellHigh}{+0.04869}
\newcommand{\RichResidualBPositiveCells}{16}
\newcommand{\AlwaysLastAccuracy}{0.830}
\newcommand{\AlwaysLastOutputRatio}{0.774}
\newcommand{\AlwaysLastFrontierAccuracy}{0.778}
\newcommand{\AlwaysLastFrontierDelta}{+0.052}
\newcommand{\AlwaysLastFrontierDeltaLow}{+0.008}
\newcommand{\AlwaysLastFrontierDeltaHigh}{+0.098}
\newcommand{\RouterAccuracy}{0.833}
\newcommand{\RouterOutputRatio}{0.885}
\newcommand{\RouterFrontierAccuracy}{0.779}
\newcommand{\RouterFrontierDelta}{+0.054}
\newcommand{\RouterFrontierDeltaLow}{+0.010}
\newcommand{\RouterFrontierDeltaHigh}{+0.103}
\newcommand{\FrontierProblems}{100}
\newcommand{\FrontierSeeds}{3}
\newcommand{\AlwaysLastSeedsAboveFrontier}{3}
\newcommand{\RouterSeedsAboveFrontier}{3}
\newcommand{\FairLossUnmasked}{+0.0128}
\newcommand{\FairLossUnmaskedLow}{-0.0033}
\newcommand{\FairLossUnmaskedHigh}{+0.0293}
\newcommand{\FairLossInBand}{-0.0016}
\newcommand{\FairLossInBandLow}{-0.0161}
\newcommand{\FairLossInBandHigh}{+0.0133}
\newcommand{\TieBCEFirst}{-0.175}
\newcommand{\TieBCERandom}{+0.095}
\newcommand{\TieListNetFirst}{+0.123}

\newcommand{\TieSeeds}{5}
\newcommand{\TieBCEFirstLow}{-0.218}
\newcommand{\TieBCEFirstHigh}{-0.137}
\newcommand{\TieBCERandomLow}{+0.050}
\newcommand{\TieBCERandomHigh}{+0.134}
\newcommand{\TieListNetLow}{+0.073}
\newcommand{\TieListNetHigh}{+0.193}
\newcommand{\GateRoutedMeanPct}{23}
\newcommand{\GateRoutedMinPct}{6}
\newcommand{\GateRoutedMaxPct}{96}
\newcommand{\EqThreeGapNWeightedPP}{+18.27}
\newcommand{\FullMinusFixedMask}{-0.0002}
\newcommand{\FullMinusFixedMaskLow}{-0.0029}
\newcommand{\FullMinusFixedMaskHigh}{+0.0019}
\newcommand{\PriorRemoval}{+0.0215}
\newcommand{\PriorRemovalLow}{+0.0032}
\newcommand{\PriorRemovalHigh}{+0.0398}
\newcommand{\ComponentControlSeeds}{10}
\newcommand{\PRMMinusTextUngated}{-0.0038}
\newcommand{\PRMMinusTextUngatedLow}{-0.0127}
\newcommand{\PRMMinusTextUngatedHigh}{+0.0053}
\newcommand{\PRMMinusTextGated}{-0.0066}
\newcommand{\PRMMinusTextGatedLow}{-0.0252}
\newcommand{\PRMMinusTextGatedHigh}{+0.0110}
\newcommand{\FeatureControlSeeds}{10}
\newcommand{\LogprobMinusText}{-0.0030}
\newcommand{\HiddenMinusText}{-0.0038}
\newcommand{\IndependentPrefixGapPP}{+5.48}
\newcommand{\IndependentPrefixGapLowPP}{+3.67}
\newcommand{\IndependentPrefixGapHighPP}{+7.59}
\newcommand{\IndependentPrefixCells}{16}
\newcommand{\IndependentPrefixExamples}{3421}
\newcommand{\IndependentPrefixGiniReliability}{0.994}
\newcommand{\SegStrideTwoPP}{+11.87}
\newcommand{\SegStrideThreePP}{+8.66}
\newcommand{\SegFixedFourPP}{+12.17}
\newcommand{\SegmentationPositiveCells}{41}
\newcommand{\TransferCode}{+0.216}
\newcommand{\TransferCodeLow}{+0.138}
\newcommand{\TransferCodeHigh}{+0.292}
\newcommand{\TransferMath}{+0.120}
\newcommand{\TransferMathLow}{+0.094}
\newcommand{\TransferMathHigh}{+0.146}
\newcommand{\TransferScience}{+0.046}
\newcommand{\TransferScienceLow}{+0.008}
\newcommand{\TransferScienceHigh}{+0.084}
\newcommand{\TransferPositiveSeeds}{15}
\newcommand{\TransferSeeds}{15}
\newcommand{\FairLossExtendedUnmasked}{+0.0184}
\newcommand{\FairLossExtendedUnmaskedLow}{+0.0040}
\newcommand{\FairLossExtendedUnmaskedHigh}{+0.0330}
\newcommand{\FairLossExtendedInBand}{+0.0022}
\newcommand{\FairLossExtendedInBandLow}{-0.0093}
\newcommand{\FairLossExtendedInBandHigh}{+0.0138}
\newcommand{\FairLossExtendedSeeds}{64}
\newcommand{\FairLossExtendedUnmaskedSeedLow}{+0.0105}
\newcommand{\FairLossExtendedUnmaskedSeedHigh}{+0.0263}
\newcommand{\FairLossExtendedInBandSeedLow}{-0.0038}
\newcommand{\FairLossExtendedInBandSeedHigh}{+0.0081}
\newcommand{\FairLossEquivalenceMargin}{0.02}
\newcommand{\MathSensitivityDelta}{+0.0261}
\newcommand{\MathSensitivityLow}{+0.0020}
\newcommand{\MathSensitivityHigh}{+0.0506}
\newcommand{\QwenMathDelta}{+0.0008}
\newcommand{\QwenMathDeltaLow}{-0.0168}
\newcommand{\QwenMathDeltaHigh}{+0.0170}
\newcommand{\TreeMinusSC}{-0.020}
\newcommand{\TreeMinusSCLow}{-0.060}
\newcommand{\TreeMinusSCHigh}{+0.017}
\newcommand{\TreeOutputRatio}{0.720}
\newcommand{\HtwoTotalTokenRatio}{0.912}
\newcommand{\HtwoAccuracyDelta}{+0.030}
\newcommand{\HtwoAccuracyDeltaLow}{-0.005}
\newcommand{\HtwoAccuracyDeltaHigh}{+0.068}
\newcommand{\HtwoWallclockRatio}{1.108}
\newcommand{\DeepSeekPRMBoNMinusSC}{-0.013}
\newcommand{\DeepSeekPRMBoNMinusSCLow}{-0.037}
\newcommand{\DeepSeekPRMBoNMinusSCHigh}{+0.013}
\newcommand{\QwenPRMBoNMinusSC}{-0.003}
\newcommand{\QwenPRMBoNMinusSCLow}{-0.037}
\newcommand{\QwenPRMBoNMinusSCHigh}{+0.023}
\newcommand{\RouterMinusUniform}{+0.127}
\newcommand{\RouterMinusUniformLow}{+0.108}
\newcommand{\RouterMinusUniformHigh}{+0.146}
\newcommand{\RouterMinusLastUngated}{-0.006}
\newcommand{\RouterMinusLastUngatedLow}{-0.016}
\newcommand{\RouterMinusLastUngatedHigh}{+0.005}
\newcommand{\RouterMinusLastGated}{+0.008}
\newcommand{\RouterMinusLastGatedLow}{-0.023}
\newcommand{\RouterMinusLastGatedHigh}{+0.037}
\newcommand{\GateEffectRouter}{+0.056}
\newcommand{\GateEffectRouterLow}{+0.020}
\newcommand{\GateEffectRouterHigh}{+0.092}
\newcommand{\GateEffectLast}{+0.042}
\newcommand{\GateEffectLastLow}{+0.005}
\newcommand{\GateEffectLastHigh}{+0.083}
\newcommand{\GateDifference}{+0.013}
\newcommand{\GateDifferenceLow}{-0.020}
\newcommand{\GateDifferenceHigh}{+0.044}
\newcommand{\FixedComposition}{+0.003}
\newcommand{\FixedCompositionLow}{-0.010}
\newcommand{\FixedCompositionHigh}{+0.015}
\newcommand{\SCKOneRatio}{0.250}
\newcommand{\SCKOneAccuracy}{0.771}
\newcommand{\SCKTwoRatio}{0.500}
\newcommand{\SCKTwoAccuracy}{0.766}
\newcommand{\SCKThreeRatio}{0.750}
\newcommand{\SCKThreeAccuracy}{0.778}
\newcommand{\SCKFourRatio}{1.000}
\newcommand{\SCKFourAccuracy}{0.780}
\newcommand{\FreshAuditCells}{5}
\newcommand{\FreshAuditExamples}{718}
\newcommand{\FreshSameDraw}{+14.25}
\newcommand{\FreshSameDrawLow}{+12.00}
\newcommand{\FreshSameDrawHigh}{+16.35}
\newcommand{\FreshStoredWinner}{+2.01}
\newcommand{\FreshStoredWinnerLow}{+1.04}
\newcommand{\FreshStoredWinnerHigh}{+3.66}
\newcommand{\FreshIndependentFour}{+4.25}
\newcommand{\FreshIndependentFourLow}{+2.51}
\newcommand{\FreshIndependentFourHigh}{+6.63}
\newcommand{\CrossfitGap}{+11.03}
\newcommand{\CrossfitGapLow}{+7.74}
\newcommand{\CrossfitGapHigh}{+14.54}
\newcommand{\CrossfitPositiveCells}{36}
\newcommand{\CrossfitAnalyzableCells}{38}
\newcommand{\CrossfitMedianGiniGap}{-0.431}
\newcommand{\CrossfitMedianGiniGapLow}{-0.677}
\newcommand{\CrossfitMedianGiniGapHigh}{-0.167}
\newcommand{\CrossfitGiniReliabilityMin}{0.989}
\newcommand{\CrossfitGiniReliabilityMax}{0.996}

\definecolor{figink}{HTML}{1F1F1F}      %
\definecolor{figinksec}{HTML}{555555}   %
\definecolor{figmuted}{HTML}{8A8A8A}    %
\definecolor{figaxis}{HTML}{7A7A7A}     %
\definecolor{figgrid}{HTML}{E6E6E6}     %
\definecolor{figfill}{HTML}{F2F2F2}     %
\definecolor{figblue}{HTML}{0072B2}     %
\definecolor{figverm}{HTML}{D55E00}     %
\colorlet{figbluetint}{figblue!9!white}
\colorlet{figvermtint}{figverm!9!white}

\providecommand{\figtick}{\fontsize{8.5}{10}\selectfont}
\providecommand{\figtext}{\fontsize{9}{10.8}\selectfont}
\providecommand{\figlabel}{\fontsize{9}{10.8}\selectfont}

\pgfplotsset{
  premier/.style={
    axis lines*=left,
    axis line style={draw=figaxis, line width=0.4pt},
    tick style={draw=figaxis, line width=0.4pt},
    tick align=outside, major tick length=2.5pt,
    ymajorgrids, grid style={draw=figgrid, line width=0.4pt},
    tick label style={font=\figtick, text=figinksec},
    label style={font=\figlabel, text=figink},
    title style={font=\figlabel, text=figink, yshift=-2pt},
    legend style={draw=none, fill=white, fill opacity=0.9, text opacity=1,
                  font=\figtick, text=figinksec, inner sep=2pt,
                  row sep=0pt, column sep=3pt},
    legend cell align=left,
    every axis plot/.append style={line join=round, line cap=round},
    scaled ticks=false,
    clip=false,
  },
}

\providecommand{\figtitle}{\fontsize{9.5}{11.4}\selectfont\bfseries}   %
\definecolor{figgraytint}{HTML}{F5F5F4}
\colorlet{figbluesoft}{figblue!16!white}
\colorlet{figvermsoft}{figverm!16!white}
\pgfplotsset{
  colormap={figseq}{color(0)=(figblue!5!white) color(0.5)=(figblue!45!white) color(1)=(figblue!92!black)},
}
\tikzset{
  figbox/.style={draw=figaxis!70, line width=0.4pt, rounded corners=3pt, fill=white,
                 align=center, font=\figtext, text=figink, inner sep=4pt},
  figboxdata/.style={figbox, fill=figgraytint},
  figboxmeasure/.style={figbox, fill=figbluetint, draw=figblue!55},
  figboxpolicy/.style={figbox, fill=figvermtint, draw=figverm!55},
  figarrow/.style={-{Latex[length=4.5pt,width=3.5pt]}, draw=figaxis, line width=0.6pt},
  figstep/.style={circle, fill=figink, text=white, font=\fontsize{7.5}{8}\selectfont\bfseries,
                  inner sep=0pt, minimum size=11pt},
  figpanellabel/.style={font=\figtitle, text=figink, anchor=south west, inner sep=0pt},
}

\tikzset{
  figcard/.style={fill=figgraytint, draw=none, rounded corners=4pt},
}

\newcommand{\MeasurementTable}{%
\begin{table}[H]
\caption{\textbf{Measurement summary.} The primary set contains
$41$ cells; the complete display contains two additional exploratory Putnam
cells with $n=23$. Pooled Gini is computed after pooling positive utility over
eligible positions within a cell. The final row is a different estimand from
Eq.~\eqref{eq:formal-gap}.}
\label{tab:measurement}
\centering
\normalsize
\setlength{\tabcolsep}{6pt}
\begin{tabular}{lcc}
\toprule
Quantity & Estimate & Coverage \\
\midrule
Formal Eq.~\eqref{eq:formal-gap} advantage & $\EqThreeGapPP$ pp & $\EqThreePositiveCells/\EqThreePrimaryCells$ positive \\
Pooled eligible-position Gini & $\PooledEligibleGini$ & $\PooledEligibleGiniHighCells/\EqThreePrimaryCells$ cells $\geq 0.80$ \\
Mean within-example eligible Gini & $\WithinExampleEligibleGini$ & $41$ cells \\
Maximum observed utility & $\MaximumObservedUtilityPP$ pp & different estimand \\
\bottomrule
\end{tabular}
\end{table}}

\newcommand{\DecisionTable}{%
\begin{table}[H]
\caption{\textbf{Separating placement from spending.} GR@10 intervals are
nominal paired split-seed bootstrap stability intervals on one frozen store,
not problem-population intervals. The router-fitted gate threshold is applied
unchanged to both placement rules. Crossing-zero intervals are reported as
no detected paired difference, not as equivalence.}
\label{tab:decisions}
\centering
\vspace{1mm}
\normalsize
\setlength{\tabcolsep}{4pt}
\begin{tabular}{lll}
\toprule
Contrast & Estimate $[95\%\ \mathrm{CI}]$ & Interpretation \\
\midrule
Router $-$ uniform, ungated & $\RouterMinusUniform\ [\RouterMinusUniformLow,\RouterMinusUniformHigh]$ & stable in-distribution diagnostic \\
Router $-$ always-last, ungated & $\RouterMinusLastUngated\ [\RouterMinusLastUngatedLow,\RouterMinusLastUngatedHigh]$ & no detected increment \\
Gate effect on router & $\GateEffectRouter\ [\GateEffectRouterLow,\GateEffectRouterHigh]$ & gate-on minus gate-off \\
Gate effect on always-last & $\GateEffectLast\ [\GateEffectLastLow,\GateEffectLastHigh]$ & gate-on minus gate-off \\
Difference in gate effects & $\GateDifference\ [\GateDifferenceLow,\GateDifferenceHigh]$ & no detected interaction \\
Router $-$ always-last, gated & $\RouterMinusLastGated\ [\RouterMinusLastGatedLow,\RouterMinusLastGatedHigh]$ & no detected increment \\
Fixed-composition accuracy audit & $\FixedComposition\ [\FixedCompositionLow,\FixedCompositionHigh]$ & problem-cluster bootstrap \\
\bottomrule
\end{tabular}
\end{table}}

\newcommand{\DepthResidualTable}{%
\begin{table}[H]
\caption{\textbf{Held-out oracle-information residual beyond a declared
21-policy depth class.} Five-fold cross-fitting, $10\%$ within-chain budget,
$\DepthResidualExamples$ records and $\DepthResidualCells$ cells; intervals resample
examples pooled across cells. This is a held-out oracle residual over the
selected depth rule, not a feasible selector's gain or a deployed-accuracy
result.}
\label{tab:depth-residual}
\centering
\small
\setlength{\tabcolsep}{4pt}
\begin{tabular}{lcc}
\toprule
Direction & Residual $[95\%\ \mathrm{CI}]$ & Positive cells \\
\midrule
A selects, B scores & $\DepthResidualA\ [\DepthResidualALow,\DepthResidualAHigh]$ & $\DepthResidualAPositiveCells/\DepthResidualCells$ \\
B selects, A scores & $\DepthResidualB\ [\DepthResidualBLow,\DepthResidualBHigh]$ & $\DepthResidualBPositiveCells/\DepthResidualCells$ \\
\bottomrule
\end{tabular}
\end{table}}

\newcommand{\RichResidualTable}{%
\begin{table}[H]
\caption{\textbf{Nested-selected enriched positional sensitivity.} The candidate
family contains the original grid and regularized length-by-rank tables. It
does not cover the full arbitrary positional class, and the selected enriched
learner trails the original grid out of sample. Intervals cluster by problems
or cells as labeled.}
\label{tab:rich-residual}
\centering
\vspace{1mm}
\normalsize
\setlength{\tabcolsep}{3.5pt}
\begin{tabular}{lcccc}
\toprule
Direction & Mean & Problem-cluster $95\%$ CI & Cell-cluster $95\%$ CI & Positive cells \\
\midrule
A selects, B scores & $\RichResidualA$ & $[\RichResidualAProblemLow,\RichResidualAProblemHigh]$ & $[\RichResidualACellLow,\RichResidualACellHigh]$ & $\RichResidualAPositiveCells/16$ \\
B selects, A scores & $\RichResidualB$ & $[\RichResidualBProblemLow,\RichResidualBProblemHigh]$ & $[\RichResidualBCellLow,\RichResidualBCellHigh]$ & $\RichResidualBPositiveCells/16$ \\
\bottomrule
\end{tabular}
\end{table}}

\newcommand{\FrontierTable}{%
\begin{table}[H]
\caption{\textbf{Exact aggregate generated-output SC frontier on
DeepSeek-R1-Distill-Qwen-14B/MATH-500.} The panel contains $\FrontierProblems$ problems and
$\FrontierSeeds$ generation seeds. Ratios use aggregate generated output only;
they do not measure total tokens or latency. CIs use paired problem bootstrap.}
\label{tab:frontier}
\centering
\vspace{1mm}
\small
\setlength{\tabcolsep}{7pt}
\begin{tabular}{lcccc}
\toprule
Policy & Acc. & Output ratio & Frontier & Delta $[95\%\ \mathrm{CI}]$ \\
\midrule
Always-last & $\AlwaysLastAccuracy$ & $\AlwaysLastOutputRatio$ & $\AlwaysLastFrontierAccuracy$ & $\AlwaysLastFrontierDelta\ [\AlwaysLastFrontierDeltaLow,\AlwaysLastFrontierDeltaHigh]$ \\
Router & $\RouterAccuracy$ & $\RouterOutputRatio$ & $\RouterFrontierAccuracy$ & $\RouterFrontierDelta\ [\RouterFrontierDeltaLow,\RouterFrontierDeltaHigh]$ \\
\bottomrule
\end{tabular}
\end{table}}

\newcommand{\IndependentScoringTable}{%
\begin{table}[H]
\caption{\textbf{Held-out scoring audits estimate different selectors.}
The five-cell audit has $\FreshAuditExamples$ exact-prefix examples. The
matrix-wide 2/2 audit covers the $\CrossfitAnalyzableCells$ primary cells that
retain
individually labeled continuation outcomes. Values are cell-equal percentage-point advantages over
the exact uniform-candidate comparator.}
\label{tab:independent-scoring}
\centering
\small
\setlength{\tabcolsep}{4pt}
\begin{tabular}{lc}
\toprule
Protocol & Gap $[95\%\ \mathrm{CI}]$ \\
\midrule
Fresh-eight same-draw selection/scoring & $\FreshSameDraw\ [\FreshSameDrawLow,\FreshSameDrawHigh]$ \\
Stored winner, eight unseen scoring draws & $\FreshStoredWinner\ [\FreshStoredWinnerLow,\FreshStoredWinnerHigh]$ \\
Fresh four select, disjoint four score & $\FreshIndependentFour\ [\FreshIndependentFourLow,\FreshIndependentFourHigh]$ \\
Exhaustive tie-neutral stored 2/2 & $\CrossfitGap\ [\CrossfitGapLow,\CrossfitGapHigh]$ \\
\bottomrule
\end{tabular}
\end{table}}

\newcommand{\RequestedComponentTable}{%
\begin{table}[H]
\caption{\textbf{Component and feature controls.} Component
contrasts use matched split seeds and GR@10. The log-probability and
hidden-state rows are matched-coverage 10-seed mean contrasts; their source
artifact did not promote a population interval. Negative or crossing-zero
results are reported as scoped nulls, not as evidence that a feature family is
generally uninformative.}
\label{tab:requested-components}
\centering
\vspace{1mm}
\small
\setlength{\tabcolsep}{4pt}
\begin{tabular}{lcl}
\toprule
Control & Estimate $[95\%\ \mathrm{CI}]$ & Scope \\
\midrule
Benchmark-conditional $-$ no support &
$\FullMinusFixedMask\ [\FullMinusFixedMaskLow,\FullMinusFixedMaskHigh]$ &
$\ComponentControlSeeds$ matched seeds \\
Remove learned depth prior &
$\PriorRemoval\ [\PriorRemovalLow,\PriorRemovalHigh]$ &
$\ComponentControlSeeds$ matched seeds \\
PRM-feature $-$ text-feature variant, ungated &
$\PRMMinusTextUngated\ [\PRMMinusTextUngatedLow,\PRMMinusTextUngatedHigh]$ &
$25$ matched seeds \\
PRM-feature $-$ text-feature variant, gated &
$\PRMMinusTextGated\ [\PRMMinusTextGatedLow,\PRMMinusTextGatedHigh]$ &
$25$ matched seeds \\
Log-probability $-$ text feature set & $\LogprobMinusText$ &
$\FeatureControlSeeds$-seed matched coverage \\
Hidden-state $-$ text feature set & $\HiddenMinusText$ &
$\FeatureControlSeeds$-seed matched coverage \\
\bottomrule
\end{tabular}
\end{table}}

\newcommand{\DomainTransferTable}{%
\begin{table}[H]
\caption{\textbf{Within-suite domain-held transfer.} The target domain is
removed from training and validation, benchmark identity is disabled, and one
shared late-only support is used. Estimates are router GR@10 over uniform with
problem-cluster intervals. They do not establish improvement over always-last
or transfer outside the benchmark suite.}
\label{tab:domain-transfer}
\centering
\small
\setlength{\tabcolsep}{5pt}
\begin{tabular}{lcc}
\toprule
Held-out domain & GR@10 over uniform $[95\%\ \mathrm{CI}]$ & Positive fits \\
\midrule
Code & $\TransferCode\ [\TransferCodeLow,\TransferCodeHigh]$ & $\TransferPositiveSeeds/\TransferSeeds$ \\
Mathematics & $\TransferMath\ [\TransferMathLow,\TransferMathHigh]$ & $\TransferPositiveSeeds/\TransferSeeds$ \\
Science & $\TransferScience\ [\TransferScienceLow,\TransferScienceHigh]$ & $\TransferPositiveSeeds/\TransferSeeds$ \\
\bottomrule
\end{tabular}
\end{table}}

\newcommand{\MeasurementRobustnessTable}{%
\begin{table}[H]
\caption{\textbf{Separately generated samples and boundary density.} The
symmetric 4/4 result uses separately generated continuation samples at
identical prefixes. Boundary thinning changes the candidate set and therefore the magnitude of
the same-draw Eq.~\eqref{eq:formal-gap} statistic; it does not test semantic
segmentation. The boundary-density sources report
point estimates only, not population intervals.}
\label{tab:measurement-robustness}
\centering
\small
\setlength{\tabcolsep}{5pt}
\begin{tabular}{lccc}
\toprule
Protocol & Cell-equal gap $[95\%\ \mathrm{CI}]$ & Coverage & Positive cells \\
\midrule
Separate A/B samples, symmetric 4/4 &
$\IndependentPrefixGapPP\ [\IndependentPrefixGapLowPP,\IndependentPrefixGapHighPP]$ pp &
$\IndependentPrefixExamples$ examples & $\IndependentPrefixCells/\IndependentPrefixCells$ \\
Every second exact boundary & $\SegStrideTwoPP$ pp & $41$ cells & $\SegmentationPositiveCells/41$ \\
Every third exact boundary & $\SegStrideThreePP$ pp & $41$ cells & $\SegmentationPositiveCells/41$ \\
Four fixed normalized anchors & $\SegFixedFourPP$ pp & $41$ cells & $\SegmentationPositiveCells/41$ \\
\bottomrule
\end{tabular}
\end{table}}

\newcommand{\LossExtensionTable}{%
\begin{table}[H]
\caption{\textbf{Fair-label loss contrast and locked seed extension.} The
24-seed comparison is primary; the 64-seed estimate is a post-lock robustness
extension, not an independent confirmation. Intervals cluster by problem.}
\label{tab:loss-extension}
\centering
\small
\setlength{\tabcolsep}{4pt}
\begin{tabular}{lcc}
\toprule
Protocol & Unmasked ListNet $-$ BCE & In benchmark-conditional support \\
\midrule
Primary, 24 seeds &
$\FairLossUnmasked\ [\FairLossUnmaskedLow,\FairLossUnmaskedHigh]$ &
$\FairLossInBand\ [\FairLossInBandLow,\FairLossInBandHigh]$ \\
Locked extension, $\FairLossExtendedSeeds$ seeds &
$\FairLossExtendedUnmasked\ [\FairLossExtendedUnmaskedLow,\FairLossExtendedUnmaskedHigh]$ &
$\FairLossExtendedInBand\ [\FairLossExtendedInBandLow,\FairLossExtendedInBandHigh]$ \\
\bottomrule
\end{tabular}
\end{table}}

\newcommand{\EfficiencyControlTable}{%
\begin{table}[H]
\caption{\textbf{Scoped inference-scaling controls.} The first two rows are a
partial fixed two-model MATH sensitivity, not the incomplete planned 2-by-2
model-by-benchmark panel. PRM best-of-$N$, the shallow tree, and the H200 assay
use their own stated cohorts and ledgers and are not pooled with the exact
DeepSeek frontier result.}
\label{tab:efficiency-controls}
\centering
\vspace{1mm}
\small
\setlength{\tabcolsep}{4pt}
\begin{tabular}{lll}
\toprule
Control & Estimate $[95\%\ \mathrm{CI}]$ & Scope \\
\midrule
Always-last over exact SC frontier &
$\MathSensitivityDelta\ [\MathSensitivityLow,\MathSensitivityHigh]$ &
cell-equal, two available MATH cells \\
Qwen3-32B/MATH-500 alone &
$\QwenMathDelta\ [\QwenMathDeltaLow,\QwenMathDeltaHigh]$ &
individually unresolved \\
PRM best-of-$N$ $-$ SC, DeepSeek &
$\DeepSeekPRMBoNMinusSC\ [\DeepSeekPRMBoNMinusSCLow,\DeepSeekPRMBoNMinusSCHigh]$ &
$100$ problems, $3$ seeds \\
PRM best-of-$N$ $-$ SC, Qwen3-32B &
$\QwenPRMBoNMinusSC\ [\QwenPRMBoNMinusSCLow,\QwenPRMBoNMinusSCHigh]$ &
$100$ problems, $3$ seeds \\
Shallow PRM tree $-$ SC &
$\TreeMinusSC\ [\TreeMinusSCLow,\TreeMinusSCHigh]$ &
$\TreeOutputRatio\times$ generated output \\
H200 always-last $-$ SC accuracy &
$\HtwoAccuracyDelta\ [\HtwoAccuracyDeltaLow,\HtwoAccuracyDeltaHigh]$ &
$\HtwoTotalTokenRatio\times$ total tokens; $\HtwoWallclockRatio\times$ time \\
\bottomrule
\end{tabular}
\end{table}}

\title{Expanding LLM Reasoning}

\author{%
  Rian Atri\\
  Keiji AI\\
  \texttt{rianatri@keiji.ai}
  \And
  Evan Luo\\
  University of California, Berkeley\\
  \texttt{evanluo25@berkeley.edu}
}

\begin{document}
\raggedbottom
\maketitle

\begin{abstract}
Extra inference compute is usually spent on sampling more reasoning chains. We
study where inside an existing chain an additional continuation should begin.
We define expansion utility, the change in correctness from restarting a chain
at a stored step, and measure it at every eligible step for nine models on six
benchmarks (41 model and benchmark cells). Restart position matters: steps
selected on one set of continuations beat uniform placement when scored on
disjoint ones, in held-out audits on 5, 16, and 38 cells ($\FreshIndependentFour$
points $[\FreshIndependentFourLow,\FreshIndependentFourHigh]$ in a fresh
five-cell audit). A fixed rule that restarts from the last eligible steps,
always-last, is a strong baseline: our learned router beats uniform placement
but shows no detected gain over it, and on DeepSeek-R1-Distill-Qwen-14B/MATH-500
always-last exceeds the exact self-consistency frontier at matched aggregate
generated output by $\AlwaysLastFrontierDelta$
$[\AlwaysLastFrontierDeltaLow,\AlwaysLastFrontierDeltaHigh]$, using
$\AlwaysLastOutputRatio\times$ the aggregate generated output of four-sample
self-consistency. Cross-fitted oracle selection still finds held-out headroom
beyond declared positional classes, a target for future selectors. Finally,
breaking step-label ties by earliest index flips the sign of a pointwise
selector's gain over uniform placement in every seed of a five-seed diagnostic
with four rollouts per step; randomized ties remove
the bias.
\end{abstract}

\section{Introduction}
\label{sec:intro}

Test-time compute is now a standard way to improve reasoning performance: models
generate longer chains of thought, sample and vote over many chains, or rerank
candidates with
verifiers~\citep{wei2022cot,wang2022selfconsistency,lightman2023lets,snell2024scaling,brown2024large}.
Most of these methods operate on complete trajectories, deciding how many to
generate and which to keep. They leave open a finer-grained question: given an existing
chain, where would one additional continuation help most?

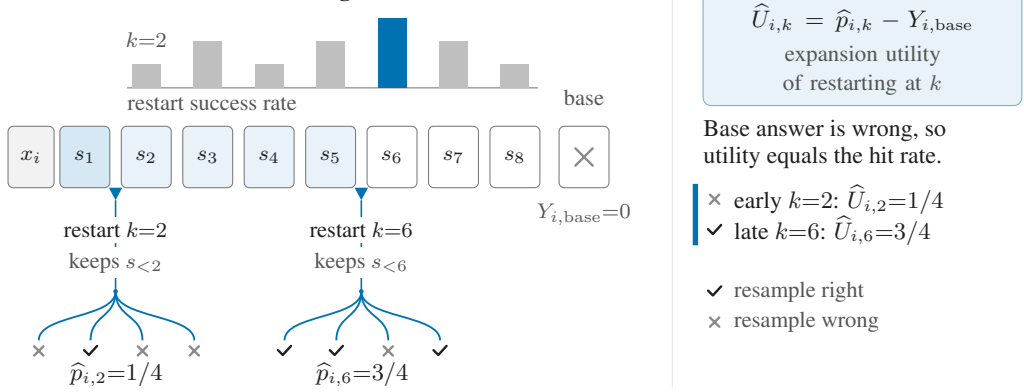
\begin{figure}[H]
\centering
\tikzset{
  schemStep/.style={draw=figaxis, line width=0.4pt,
    rounded corners=2pt, minimum width=0.26in, minimum height=0.32in,
    inner sep=0pt, font=\figtick, text=figink},
  schemCheck/.pic={\draw[figink, line width=0.9pt, line cap=round, line join=round]
    (-0.034,0.004) -- (-0.009,-0.021) -- (0.034,0.026);},
  schemCross/.pic={\draw[figmuted, line width=0.9pt, line cap=round]
    (-0.024,-0.024) -- (0.024,0.024) (-0.024,0.024) -- (0.024,-0.024);},
}%
\begin{tikzpicture}[x=1in, y=1in, >=Latex]
\node[anchor=west, font=\figtitle, text=figink, inner sep=0pt] at (0.06,2.17)
  {Where a chain is restarted changes whether it succeeds};
\draw[figaxis, line width=0.4pt] (0.74,1.66) -- (2.88,1.66);
\fill[figmuted!55!white] (0.765,1.66) rectangle (0.915,1.78);
\fill[figmuted!55!white] (1.085,1.66) rectangle (1.235,1.90);
\fill[figmuted!55!white] (1.405,1.66) rectangle (1.555,1.78);
\fill[figmuted!55!white] (1.725,1.66) rectangle (1.875,1.90);
\fill[figblue]  (2.045,1.66) rectangle (2.195,2.02);
\fill[figmuted!55!white] (2.365,1.66) rectangle (2.515,1.90);
\fill[figmuted!55!white] (2.685,1.66) rectangle (2.835,1.78);
\node[anchor=south, font=\figtick, text=figinksec, inner sep=0pt] at (0.84,1.87) {$k{=}2$};
\node[anchor=west, font=\figtick, text=figinksec, inner sep=0pt] at (0.74,1.58)
  {restart success rate};
\node[schemStep, minimum width=0.24in, fill=figfill] at (0.24,1.30) {$x_i$};
\node[schemStep, fill=figbluesoft] at (0.52,1.30) {$s_1$};
\node[schemStep, fill=figbluetint] at (0.84,1.30) {$s_2$};
\node[schemStep, fill=figbluetint] at (1.16,1.30) {$s_3$};
\node[schemStep, fill=figbluetint] at (1.48,1.30) {$s_4$};
\node[schemStep, fill=figbluetint] at (1.80,1.30) {$s_5$};
\node[schemStep, fill=white] at (2.12,1.30) {$s_6$};
\node[schemStep, fill=white] at (2.44,1.30) {$s_7$};
\node[schemStep, fill=white] at (2.76,1.30) {$s_8$};
\node[schemStep, fill=white] at (3.12,1.30) {};
\draw[figmuted, line width=0.9pt, line cap=round]
  (3.075,1.255) -- (3.165,1.345) (3.075,1.345) -- (3.165,1.255);
\node[anchor=south, font=\figtick, text=figinksec, inner sep=0pt] at (3.12,1.565) {base};
\node[anchor=north, font=\figtick, text=figinksec, inner sep=0pt] at (3.12,1.06)
  {$Y_{i,\mathrm{base}}{=}0$};
\fill[figblue] (0.645,1.14) -- (0.715,1.14) -- (0.68,1.075) -- cycle;
\fill[figblue] (1.925,1.14) -- (1.995,1.14) -- (1.96,1.075) -- cycle;
\draw[figblue, line width=0.6pt] (0.68,1.075) -- (0.68,0.60);
\draw[figblue, line width=0.6pt] (1.96,1.075) -- (1.96,0.60);
\fill[figblue] (0.68,0.60) circle (0.012);
\fill[figblue] (1.96,0.60) circle (0.012);
\node[font=\figtick, text=figink, inner sep=2pt, fill=white] at (0.68,0.92)
  {restart $k{=}2$};
\node[font=\figtick, text=figinksec, inner sep=2pt, fill=white] at (0.68,0.75)
  {keeps $s_{<2}$};
\node[font=\figtick, text=figink, inner sep=2pt, fill=white] at (1.96,0.92)
  {restart $k{=}6$};
\node[font=\figtick, text=figinksec, inner sep=2pt, fill=white] at (1.96,0.75)
  {keeps $s_{<6}$};
\draw[figblue, line width=0.7pt] (0.68,0.60) .. controls (0.68,0.46) and (0.28,0.48) .. (0.28,0.345);
\draw[figblue, line width=0.7pt] (0.68,0.60) .. controls (0.68,0.46) and (0.55,0.48) .. (0.55,0.345);
\draw[figblue, line width=0.7pt] (0.68,0.60) .. controls (0.68,0.46) and (0.82,0.48) .. (0.82,0.345);
\draw[figblue, line width=0.7pt] (0.68,0.60) .. controls (0.68,0.46) and (1.09,0.48) .. (1.09,0.345);
\draw[figblue, line width=0.7pt] (1.96,0.60) .. controls (1.96,0.46) and (1.56,0.48) .. (1.56,0.345);
\draw[figblue, line width=0.7pt] (1.96,0.60) .. controls (1.96,0.46) and (1.83,0.48) .. (1.83,0.345);
\draw[figblue, line width=0.7pt] (1.96,0.60) .. controls (1.96,0.46) and (2.10,0.48) .. (2.10,0.345);
\draw[figblue, line width=0.7pt] (1.96,0.60) .. controls (1.96,0.46) and (2.37,0.48) .. (2.37,0.345);
\pic at (0.28,0.29) {schemCross};
\pic at (0.55,0.29) {schemCheck};
\pic at (0.82,0.29) {schemCross};
\pic at (1.09,0.29) {schemCross};
\pic at (1.56,0.29) {schemCheck};
\pic at (1.83,0.29) {schemCheck};
\pic at (2.10,0.29) {schemCross};
\pic at (2.37,0.29) {schemCheck};
\node[font=\figlabel\bfseries, text=figink, inner sep=0pt] at (0.685,0.17)
  {$\widehat p_{i,2}{=}1/4$};
\node[font=\figlabel\bfseries, text=figink, inner sep=0pt] at (1.965,0.17)
  {$\widehat p_{i,6}{=}3/4$};
\draw[figgrid, line width=0.4pt] (3.58,0.10) -- (3.58,2.18);
\node[figboxmeasure, minimum width=1.66in, text width=1.50in] at (4.57,1.86)
  {$\widehat U_{i,k}=\widehat p_{i,k}-Y_{i,\mathrm{base}}$\\[2pt]
   {\figtick\color{figinksec}expansion utility of restarting at $k$}};
\node[anchor=west, font=\figtext, text=figink, inner sep=0pt] at (3.74,1.44)
  {Base answer is wrong, so};
\node[anchor=west, font=\figtext, text=figink, inner sep=0pt] at (3.74,1.30)
  {utility equals the hit rate.};
\draw[figblue, line width=2pt] (3.70,0.84) -- (3.70,1.16);
\pic at (3.80,1.08) {schemCross};
\node[anchor=west, font=\figtext, text=figink, inner sep=0pt] at (3.90,1.08)
  {early $k{=}2$: $\widehat U_{i,2}{=}1/4$};
\pic at (3.80,0.92) {schemCheck};
\node[anchor=west, font=\figtext, text=figink, inner sep=0pt] at (3.90,0.92)
  {late $k{=}6$: $\widehat U_{i,6}{=}3/4$};
\pic at (3.80,0.60) {schemCheck};
\node[anchor=west, font=\figtick, text=figinksec, inner sep=0pt] at (3.90,0.60)
  {resample right};
\pic at (3.80,0.44) {schemCross};
\node[anchor=west, font=\figtick, text=figinksec, inner sep=0pt] at (3.90,0.44)
  {resample wrong};
\useasboundingbox (0,0.02) rectangle (5.5,2.30);
\end{tikzpicture}%
\caption{\textbf{Expansion utility.} Restarting a stored chain at step $k$ keeps
the prefix $s_{<k}$ and samples $n$ continuations. Their success rate
$\widehat p_{i,k}$, taken over all eligible steps, forms the chain's restart map,
and $\widehat U_{i,k}=\widehat p_{i,k}-Y_{i,\mathrm{base}}$ is the expansion
utility of restarting at $k$. In this schematic an early restart rarely
succeeds and a late one usually does.}
\label{fig:measurement-schematic}
\end{figure}

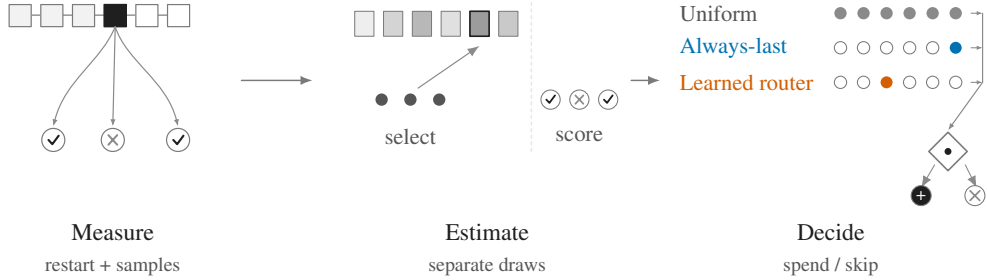
\begin{figure}[H]
\centering
\tikzset{
  wfStep/.style={draw=figaxis, line width=0.45pt},
  wfSub/.style={font=\fontsize{7.5}{9}\selectfont, text=figinksec, inner sep=0pt},
  wfMini/.style={font=\figtick, text=figinksec, inner sep=0pt},
}
\begin{tikzpicture}[x=1in,y=1in,>=Latex,line cap=round,line join=round]
\begin{scope}[xshift=-0.108in]
\draw[draw=figaxis, line width=0.45pt] (0.36,1.42) -- (1.185,1.42);
\foreach \x in {0.30,0.465,0.63} {
  \fill[figfill] (\x,1.36) rectangle +(0.115,0.115);
  \draw[wfStep] (\x,1.36) rectangle +(0.115,0.115);
}
\fill[figink] (0.795,1.36) rectangle +(0.115,0.115);
\draw[draw=figink, line width=0.55pt] (0.795,1.36) rectangle +(0.115,0.115);
\foreach \x in {0.96,1.125} {
  \fill[white] (\x,1.36) rectangle +(0.115,0.115);
  \draw[wfStep] (\x,1.36) rectangle +(0.115,0.115);
}
\draw[draw=figmuted, line width=0.45pt, -{Latex[length=2.7pt,width=2pt]}]
  (0.8525,1.36) .. controls (0.78,1.18) and (0.59,1.08) .. (0.53,0.84);
\draw[draw=figmuted, line width=0.45pt, -{Latex[length=2.7pt,width=2pt]}]
  (0.8525,1.36) -- (0.84,0.84);
\draw[draw=figmuted, line width=0.45pt, -{Latex[length=2.7pt,width=2pt]}]
  (0.8525,1.36) .. controls (0.91,1.18) and (1.12,1.08) .. (1.18,0.84);
\foreach \x in {0.53,0.84,1.18} {
  \draw[draw=figaxis, line width=0.45pt, fill=white] (\x,0.77) circle (0.06);
}
\draw[draw=figink, line width=0.8pt]
  (0.505,0.77) -- (0.525,0.748) -- (0.558,0.792);
\draw[draw=figmuted, line width=0.75pt]
  (0.817,0.795) -- (0.863,0.745) (0.817,0.745) -- (0.863,0.795);
\draw[draw=figink, line width=0.8pt]
  (1.155,0.77) -- (1.175,0.748) -- (1.208,0.792);

\fill[figinksec!10!white] (2.10,1.31) rectangle +(0.10,0.13);
\fill[figinksec!25!white] (2.25,1.31) rectangle +(0.10,0.13);
\fill[figinksec!42!white] (2.40,1.31) rectangle +(0.10,0.13);
\fill[figinksec!18!white] (2.55,1.31) rectangle +(0.10,0.13);
\fill[figinksec!58!white] (2.70,1.31) rectangle +(0.10,0.13);
\fill[figinksec!32!white] (2.85,1.31) rectangle +(0.10,0.13);
\foreach \x in {2.10,2.25,2.40,2.55,2.70,2.85} {
  \draw[draw=figaxis, line width=0.35pt] (\x,1.31) rectangle +(0.10,0.13);
}
\draw[draw=figink, line width=0.6pt] (2.70,1.31) rectangle +(0.10,0.13);
\draw[figarrow, draw=figmuted, line width=0.45pt]
  (2.43,1.04) -- (2.75,1.26);
\foreach \x in {2.24,2.39,2.54} {
  \fill[figinksec] (\x,0.98) circle (0.032);
}
\draw[draw=figgrid, line width=0.55pt, dash pattern=on 1.1pt off 1.4pt]
  (3.02,0.71) -- (3.02,1.49);
\foreach \x in {3.12,3.27,3.42} {
\draw[draw=figaxis, line width=0.4pt, fill=white] (\x,0.98) circle (0.05);
}
\draw[draw=figink, line width=0.7pt] (3.098,0.98) -- (3.114,0.963) -- (3.143,1.002);
\draw[draw=figmuted, line width=0.65pt]
  (3.249,1.002) -- (3.291,0.958) (3.249,0.958) -- (3.291,1.002);
\draw[draw=figink, line width=0.7pt] (3.398,0.98) -- (3.414,0.963) -- (3.443,1.002);
\node[wfMini] at (2.39,0.79) {select};
\node[wfMini] at (3.27,0.79) {score};

\begin{scope}[xshift=-0.15in]
\node[anchor=west, font=\figtick, text=figinksec, inner sep=0pt] at (3.94,1.43) {Uniform};
\node[anchor=west, font=\figtick, text=figblue, inner sep=0pt] at (3.94,1.25) {Always-last};
\node[anchor=west, font=\figtick, text=figverm, inner sep=0pt] at (3.94,1.07) {Learned router};
\foreach \x in {4.78,4.90,5.02,5.14,5.26,5.38} {
  \fill[figmuted] (\x,1.43) circle (0.032);
  \draw[draw=figaxis, line width=0.4pt, fill=white] (\x,1.25) circle (0.032);
  \draw[draw=figaxis, line width=0.4pt, fill=white] (\x,1.07) circle (0.032);
}
\fill[figblue] (5.38,1.25) circle (0.036);
\draw[draw=white, line width=0.7pt] (5.38,1.25) circle (0.036);
\fill[figverm] (5.02,1.07) circle (0.036);
\draw[draw=white, line width=0.7pt] (5.02,1.07) circle (0.036);
\foreach \y in {1.43,1.25,1.07} {
  \draw[draw=figmuted, line width=0.4pt, -{Latex[length=2.2pt,width=1.7pt]}]
    (5.46,\y) -- (5.52,\y);
}
\draw[draw=figmuted, line width=0.45pt] (5.52,1.43) -- (5.52,1.07);
\draw[draw=figmuted, line width=0.45pt, -{Latex[length=2.7pt,width=2pt]}]
  (5.52,1.07) -- (5.34,0.82);
\draw[draw=figaxis, line width=0.6pt, fill=white]
  (5.34,0.81) -- (5.44,0.71) -- (5.34,0.61) -- (5.24,0.71) -- cycle;
\fill[figink] (5.34,0.71) circle (0.018);
\draw[figarrow, draw=figmuted, line width=0.4pt]
  (5.27,0.64) -- (5.20,0.52);
\draw[figarrow, draw=figmuted, line width=0.4pt]
  (5.43,0.64) -- (5.49,0.52);
\draw[draw=figaxis, line width=0.4pt, fill=figink] (5.20,0.48) circle (0.052);
\node[font=\fontsize{6.6}{7}\selectfont, text=white, inner sep=0pt] at (5.20,0.48) {+};
\draw[draw=figaxis, line width=0.4pt, fill=white] (5.48,0.48) circle (0.052);
\draw[draw=figmuted, line width=0.65pt]
  (5.455,0.505) -- (5.505,0.455) (5.455,0.455) -- (5.505,0.505);
\end{scope}

\draw[figarrow, draw=figaxis, line width=0.55pt] (1.51,1.08) -- (1.88,1.08);
\draw[figarrow, draw=figaxis, line width=0.55pt] (3.54,1.08) -- (3.69,1.08);

\node[anchor=center, font=\figlabel, text=figink, inner sep=0pt] at (0.84,0.29) {Measure};
\node[wfSub] at (0.84,0.11) {restart + samples};
\node[anchor=center, font=\figlabel, text=figink, inner sep=0pt] at (2.79,0.29) {Estimate};
\node[wfSub] at (2.79,0.11) {separate draws};
\node[anchor=center, font=\figlabel, text=figink, inner sep=0pt] at (4.58,0.29) {Decide};
\node[wfSub] at (4.58,0.11) {spend / skip};
\end{scope}

\pgfresetboundingbox
\path[use as bounding box, draw=none, line width=0pt] (0,0) rectangle (5.466,1.70);
\end{tikzpicture}%
\caption{\textbf{Study design.} We restart a stored chain at every eligible step
and sample continuations (measure); estimate opportunity from the resulting
restart map, both on the same draws and on held-out draws that separate
selection from scoring (estimate); and compare placement rules (uniform,
always-last, and a learned router), with a confidence gate that decides whether
to spend a restart at all (decide).}
\label{fig:framework}
\end{figure}

We define the correctness gain from restarting a chain at a stored step as its
\emph{expansion utility} (Figure~\ref{fig:measurement-schematic}). Estimating
it requires intervening at every eligible step and observing outcomes, rather
than inferring step importance from a verifier score. The resulting restart
maps support four distinct measurements (Figure~\ref{fig:framework}): (i) opportunity when the same draws select and
score a step; (ii) whether the selected steps keep their advantage under held-out scoring; (iii)
oracle headroom beyond declared positional classes; and (iv) what a feasible
selector captures without observing outcomes.

We measure expansion utility at every eligible step for nine models (8B to
123B parameters) on six benchmarks. Each model and benchmark pair is a
\emph{cell}, and 41 cells form the primary set. Restarting from each chain's
top-ranked steps at a 10\% budget (one or two steps per chain) exceeds the mean
over eligible steps by $\EqThreeGapPP\pp$ when the same draws select and score
(Figure~\ref{fig:atlas}). This same-draw statistic is optimistic by
construction, so we test the selection on held-out draws: in audits on 5, 16, and
38 cells, positions selected on one set of continuations still beat uniform
placement when scored on disjoint ones (Figure~\ref{fig:ladder}).

Across $\RouterUniformWins$ of $\RouterUniformSeeds$ split-and-refit runs on a
frozen dataset, our learned router exceeds uniform placement in GR@10 but shows
no detected gain over always-last, a fixed rule that restarts from the last
eligible stored steps and has no learned component (Figure~\ref{fig:decisions}a). On DeepSeek-R1-Distill-Qwen-14B/MATH-500,
always-last also clears the exact self-consistency frontier at
$\AlwaysLastOutputRatio\times$ the aggregate generated output of four-sample
self-consistency, SC(4) (Figure~\ref{fig:frontier}); in the one other
available cell, the margin is unresolved.

Cross-fitted oracle selection still finds held-out headroom beyond the declared
positional classes (Figure~\ref{fig:decisions}b), a target for selectors that
go beyond them. Finally, we identify a pitfall for
step-level supervision: when several steps tie for the positive label, breaking
ties by position can flip the sign of a pointwise selector's gain. Randomized
tie breaking removes the systematic early-position preference, and the
evaluated ListNet loss is invariant to tie resolution
(Figure~\ref{fig:tie-correction}).

\SectionSpace
\section{Expansion Utility}
\label{sec:framework}

Let $x_i$ be a problem and let $s_{i,0:K_i}$ be a model-generated reasoning
chain segmented at stored step boundaries. Restarting at position $k$ keeps the
prefix $s_{i,<k}$ fixed and samples a new suffix. Each chain is stored as at
most $12$ segments, and a restart regenerates all text after its prefix. With
correctness outcome $Y_{i,k,j}\in\{0,1\}$ for continuation draw $j$, the
observed restart rate and expansion utility are
\begin{equation}
  \widehat p_{i,k}=\frac{1}{n}\sum_{j=1}^{n}Y_{i,k,j},
  \qquad
  \widehat U_{i,k}=\widehat p_{i,k}-Y_{i,\mathrm{base}},
  \label{eq:utility}
\end{equation}
where $Y_{i,\mathrm{base}}\in\{0,1\}$ is the correctness of the original chain.
Thus $\widehat p_{i,k}$ is the empirical success rate of continuations from
step $k$, and $\widehat U_{i,k}$ is its gain over the base outcome. The primary
sweep uses $n=4$ (two or eight in three cells;
Appendix~\ref{sec:reproducibility}). The eligible set $\Eset_i$ contains
restart positions $1,\ldots,K_i-1$; index 0, which would regenerate the whole
chain, is excluded. At a nominal budget $b=0.10$, $m_i=\lceil b|\Eset_i|\rceil$, which is one
step for chains with at most ten eligible positions and two for the
eleven-position chains produced by the 12-segment cap. For the $N$ included chains of a cell, the statistic compares
each chain's top-ranked steps with the mean over its eligible steps:
\begin{equation}
  \widehat\Delta
  =\frac{1}{N}\sum_{i=1}^{N}\left[
  \frac{1}{m_i}\sum_{k\in\Top_{m_i}(\widehat p_i)}\widehat p_{i,k}
  -\frac{1}{|\Eset_i|}\sum_{k\in\Eset_i}\widehat p_{i,k}
  \right].
  \label{eq:formal-gap}
\end{equation}
Because the same continuations select and score the chosen positions,
Eq.~\eqref{eq:formal-gap} is a \emph{same-draw} statistic: it measures
opportunity, not what a feasible selector gains. To test whether the selection
replicates, we also select positions on one set of continuation outcomes and
score them on disjoint outcomes. These held-out protocols differ in how many
draws select and score, so we report each separately rather than using one to
correct another. We summarize concentration with two Gini coefficients
(higher means more concentrated) that answer different questions: a pooled
Gini, which clips negative utility and pools positive utility over all
eligible positions in a cell, and a within-chain Gini, computed per chain and
then averaged.

\SectionSpace
\section{Experimental Setup}
\label{sec:implementation}

\paragraph{Models and tasks}
We evaluate nine models from 8B to 123B parameters on
MATH-500~\citep{hendrycks2021math,lightman2023lets}, GPQA~\citep{rein2024gpqa},
HumanEval~\citep{chen2021evaluating}, AIME 2025, HMMT, and MBPP, with Putnam as
an exploratory
benchmark~\citep{aime2025,hmmt2024,austin2021program,putnam2024}. The primary
set contains 41 nondegenerate model and benchmark cells with 30 to 500 problems
each; not every model is run on every benchmark, and the two exploratory Putnam cells are reported separately and
excluded from aggregates (Appendix~\ref{sec:correction-record}). Checkpoint identifiers, graders, and
generation settings are given in Appendix~\ref{sec:reproducibility}, and
per-cell counts in Table~\ref{tab:all-cells}.

\paragraph{Uncertainty}
We report coverage as cell counts. Held-out scoring
intervals use a crossed percentile bootstrap over shared problems and cells.
Oracle-residual intervals resample examples pooled across cells for the depth
class and problem clusters for the enriched class.
Intervals for the router comparisons (Section~\ref{sec:policy-boundary})
resample paired split seeds and describe stability across splits of one frozen
dataset, not problem-population uncertainty. The self-consistency comparison
uses a paired problem bootstrap that keeps every seed of a resampled problem
together. Model fitting, calibration, and held-out scoring follow frozen
protocols.

\SectionSpace
\section{Restart Opportunity Is Concentrated and Survives Held-Out Scoring}
\label{sec:measurement}

\begin{figure}[H]
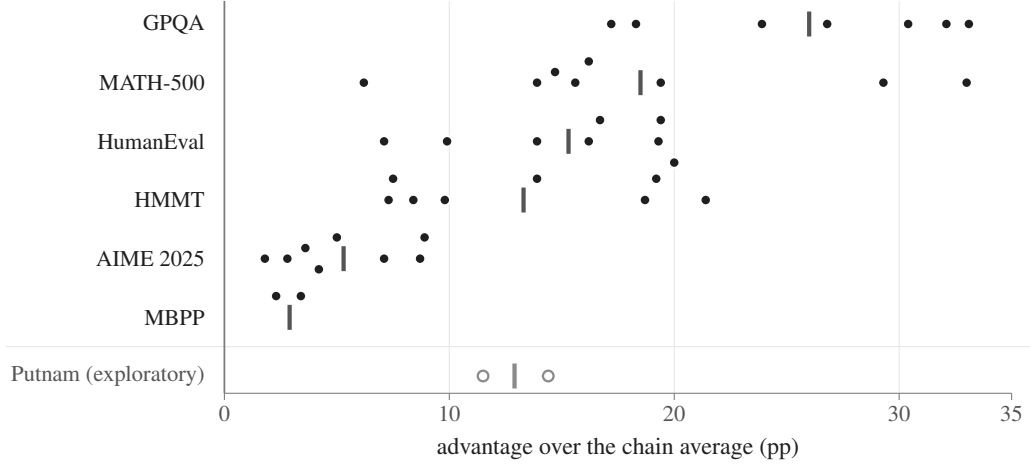

\centering
\IfFileExists{figs/data/atlas_cells.tex}{\input{figs/data/atlas_cells.tex}}{\input{data/atlas_cells.tex}}%
\begin{tikzpicture}[x=1in,y=1in]
\path[use as bounding box] (0,-0.12) rectangle (5.50,2.31);
\def\axleft{1.20}%
\def\axright{5.30}%
\def\axbottom{0.25}%
\pgfmathsetmacro{\putnamy}{\AtlasFirstRowYIn}
\pgfmathsetmacro{\primbase}{\AtlasFirstRowYIn+\AtlasPutnamGapPt/72}
\pgfmathsetmacro{\axtop}{\primbase+(\AtlasPrimaryBenchmarkCount-1)*\AtlasRowPitchPt/72+\AtlasPutnamBandHalfIn}
\pgfmathsetmacro{\sepy}{\AtlasFirstRowYIn+\AtlasPutnamGapPt/144}
\draw[draw=figgrid,line width=0.4pt] (0.06,\sepy) -- (5.43,\sepy);

\foreach \vv in {10,20,30} {
  \pgfmathsetmacro{\gx}{\axleft+(\axright-\axleft)*\vv/35}
  \draw[draw=figgrid,line width=0.4pt] (\gx,\axbottom) -- (\gx,\axtop);
}
\draw[draw=figaxis,line width=0.65pt] (\axleft,\axbottom) -- (\axleft,\axtop);
\draw[draw=figaxis,line width=0.4pt] (\axleft,\axbottom) -- (\axright,\axbottom);

\foreach \rr in {1,...,\AtlasPrimaryBenchmarkCount} {
  \edef\bb{\csname AtlasBenchKey@\rr\endcsname}
  \edef\blabel{\csname AtlasBenchLabel@\rr\endcsname}
  \edef\bmean{\csname AtlasBenchMean@\rr\endcsname}
  \pgfmathsetmacro{\rowy}{\primbase+(\rr-1)*\AtlasRowPitchPt/72}
  \node[anchor=east,font=\figtick,text=figink,inner sep=0pt]
    at (1.10,\rowy) {\blabel};
  \pgfmathsetmacro{\meanx}{\axleft+(\axright-\axleft)*\bmean/35}
  \draw[draw=figinksec,line width=1.6pt]
    (\meanx,\rowy-0.065) -- (\meanx,\rowy+0.065);
  \edef\cellmodels{\csname AtlasModels@\bb\endcsname}
  \foreach \mm in \cellmodels {
    \ifcsname AtlasVal@\mm @\bb\endcsname
      \edef\cellval{\csname AtlasVal@\mm @\bb\endcsname}
      \edef\jitter{\csname AtlasJitter@\mm @\bb\endcsname}
      \pgfmathsetmacro{\dotx}{\axleft+(\axright-\axleft)*\cellval/35}
      \filldraw[fill=figink,draw=white,line width=0.8pt]
        ([yshift=\jitter] \dotx,\rowy) circle[radius=2.0pt];
    \fi
  }
}

\node[anchor=east,font=\figtick,text=figinksec,inner sep=0pt]
  at (1.10,\putnamy) {Putnam (exploratory)};
\pgfmathsetmacro{\explmeanx}{\axleft+(\axright-\axleft)*\AtlasExploratoryMean/35}
\draw[draw=figmuted,line width=1.6pt]
  (\explmeanx,\putnamy-0.065) -- (\explmeanx,\putnamy+0.065);
\edef\explbench{putnam}
\edef\cellmodels{\csname AtlasModels@\explbench\endcsname}
\foreach \mm in \cellmodels {
  \ifcsname AtlasVal@\mm @\explbench\endcsname
    \edef\cellval{\csname AtlasVal@\mm @\explbench\endcsname}
    \pgfmathsetmacro{\dotx}{\axleft+(\axright-\axleft)*\cellval/35}
    \filldraw[fill=white,draw=figmuted,line width=0.85pt]
      (\dotx,\putnamy) circle[radius=2.0pt];
  \fi
}

\foreach \vv in {0,10,20,30,35} {
  \pgfmathsetmacro{\tx}{\axleft+(\axright-\axleft)*\vv/35}
  \draw[draw=figaxis,line width=0.4pt] (\tx,\axbottom) -- (\tx,\axbottom-0.035);
  \node[anchor=base,font=\figtick,text=figinksec,inner sep=0pt]
    at (\tx,0.105) {\vv};
}
\node[anchor=center,font=\figlabel,text=figink,inner sep=0pt]
  at (3.25,-0.04) {advantage over the chain average (pp)};
\end{tikzpicture}%
\caption{\textbf{Same-draw restart opportunity varies widely across cells.}
Each dot is one model and benchmark cell: the Eq.~\eqref{eq:formal-gap}
advantage of the top-ranked steps at a $10\%$ budget over the chain's mean
eligible step, in percentage points; ticks mark benchmark means. The statistic
is non-negative by construction; Figure~\ref{fig:ladder} tests whether the
selection replicates on held-out draws. The two exploratory Putnam cells
(hollow, $n=23$) are excluded from aggregates. Per-cell values are
in Appendix Table~\ref{tab:all-cells}.}
\label{fig:atlas}
\end{figure}

Same-draw opportunity averages $\EqThreeGapPP\pp$ across the
$\EqThreePrimaryCells$ primary cells and varies substantially across models and
benchmarks (Figure~\ref{fig:atlas}). Because a chain's top-ranked mean cannot
fall below its eligible mean, the sign of this statistic is not informative;
its magnitude, its concentration, and its survival under held-out scoring
are. For
utilities clipped at zero, positive restart value is highly concentrated within
cells (pooled Gini $\PooledEligibleGini$ on average, at least $0.80$ in
$\PooledEligibleGiniHighCells$ of 41 cells) and uneven within chains (mean
within-chain Gini $\WithinExampleEligibleGini$ among chains with nonzero
clipped utility). Cells with more concentrated positive utility
have smaller advantages ($r=\GiniGapCorrelation$ across the 41 primary cells
of Appendix Table~\ref{tab:all-cells}).

\begin{figure}[H]
\centering
\begin{tikzpicture}
\begin{axis}[
  premier,
  at={(207.00pt,0.36in)}, anchor=south west,
  width=1.32in, height=1.76in, scale only axis,
  xmin=0, xmax=18,
  ymin=0.5, ymax=6.5,
  xtick={0,5,10,15},
  ytick=\empty,
  xlabel={advantage over uniform (pp)},
  xlabel style={font=\figlabel, text=figinksec},
]
\draw[draw=none, fill=figgraytint]
  (axis cs:0,0.5) rectangle (axis cs:18,4.5);
\draw[draw=figgrid, line width=0.4pt]
  (axis cs:5,0.5) -- (axis cs:5,6.5);
\draw[draw=figgrid, line width=0.4pt]
  (axis cs:10,0.5) -- (axis cs:10,6.5);
\draw[draw=figgrid, line width=0.4pt]
  (axis cs:15,0.5) -- (axis cs:15,6.5);
\draw[draw=figaxis, line width=0.5pt]
  (axis cs:0,0.5) -- (axis cs:0,6.5);
\draw[draw=figmuted, line width=0.9pt]
  (axis cs:\FreshSameDrawLow,5) -- (axis cs:\FreshSameDrawHigh,5);
\draw[draw=figblue, line width=0.9pt]
  (axis cs:\CrossfitGapLow,4) -- (axis cs:\CrossfitGapHigh,4);
\draw[draw=figblue, line width=0.9pt]
  (axis cs:\IndependentPrefixGapLowPP,3) -- (axis cs:\IndependentPrefixGapHighPP,3);
\draw[draw=figblue, line width=0.9pt]
  (axis cs:\FreshIndependentFourLow,2) -- (axis cs:\FreshIndependentFourHigh,2);
\draw[draw=figblue, line width=0.9pt]
  (axis cs:\FreshStoredWinnerLow,1) -- (axis cs:\FreshStoredWinnerHigh,1);
\addplot[only marks, mark=*, mark options={draw=white, line width=0.8pt, fill=figmuted},
  mark size=2.6pt, forget plot]
  coordinates {(\EqThreeGapPP,6) (\FreshSameDraw,5)};
\addplot[only marks, mark=*, mark options={draw=white, line width=0.8pt, fill=figblue},
  mark size=2.6pt, forget plot]
  coordinates {(\CrossfitGap,4) (\IndependentPrefixGapPP,3)
    (\FreshIndependentFour,2) (\FreshStoredWinner,1)};
\node[anchor=east, font=\figtick, text=figink, inner sep=0pt]
  at ([xshift=-6pt] axis cs:0,6) {Eq. (2), all \EqThreePrimaryCells{} primary cells};
\node[anchor=east, font=\figtick, text=figink, inner sep=0pt]
  at ([xshift=-6pt] axis cs:0,5) {Fresh 8 draws, \FreshAuditCells{} cells};
\node[anchor=east, font=\figtick, text=figink, inner sep=0pt]
  at ([xshift=-6pt] axis cs:0,4) {Stored 2/2 split, \CrossfitAnalyzableCells{} cells};
\node[anchor=east, font=\figtick, text=figink, inner sep=0pt]
  at ([xshift=-6pt] axis cs:0,3) {Separate A/B samples, \IndependentPrefixCells{} cells};
\node[anchor=east, font=\figtick, text=figink, inner sep=0pt]
  at ([xshift=-6pt] axis cs:0,2) {Fresh 4 select, 4 score, \FreshAuditCells{} cells};
\node[anchor=east, font=\figtick, text=figink, inner sep=0pt]
  at ([xshift=-6pt] axis cs:0,1) {Stored winner, 8 unseen draws, \FreshAuditCells{} cells};
\node[anchor=west, font=\figtick, text=figinksec, inner sep=0pt]
  at ([xshift=3.4pt] axis cs:18,6) {$\EqThreeGapPP$};
\node[anchor=west, font=\figtick, text=figinksec, inner sep=0pt]
  at ([xshift=3.4pt] axis cs:18,5) {$\FreshSameDraw\ [\FreshSameDrawLow,\ \FreshSameDrawHigh]$};
\node[anchor=west, font=\figtick, text=figinksec, inner sep=0pt]
  at ([xshift=3.4pt] axis cs:18,4) {$\CrossfitGap\ [\CrossfitGapLow,\ \CrossfitGapHigh]$};
\node[anchor=west, font=\figtick, text=figinksec, inner sep=0pt]
  at ([xshift=3.4pt] axis cs:18,3) {$\IndependentPrefixGapPP\ [\IndependentPrefixGapLowPP,\ \IndependentPrefixGapHighPP]$};
\node[anchor=west, font=\figtick, text=figinksec, inner sep=0pt]
  at ([xshift=3.4pt] axis cs:18,2) {$\FreshIndependentFour\ [\FreshIndependentFourLow,\ \FreshIndependentFourHigh]$};
\node[anchor=west, font=\figtick, text=figinksec, inner sep=0pt]
  at ([xshift=3.4pt] axis cs:18,1) {$\FreshStoredWinner\ [\FreshStoredWinnerLow,\ \FreshStoredWinnerHigh]$};
\draw[draw=figaxis, line width=0.4pt]
  ([xshift=-144.31pt] axis cs:0,6.3) -- ([xshift=-144.31pt] axis cs:0,4.7);
\draw[draw=figaxis, line width=0.4pt]
  ([xshift=-144.31pt] axis cs:0,6.3) -- ([xshift=-140.31pt] axis cs:0,6.3);
\draw[draw=figaxis, line width=0.4pt]
  ([xshift=-144.31pt] axis cs:0,4.7) -- ([xshift=-140.31pt] axis cs:0,4.7);
\draw[draw=figaxis, line width=0.4pt]
  ([xshift=-144.31pt] axis cs:0,4.3) -- ([xshift=-144.31pt] axis cs:0,0.7);
\draw[draw=figaxis, line width=0.4pt]
  ([xshift=-144.31pt] axis cs:0,4.3) -- ([xshift=-140.31pt] axis cs:0,4.3);
\draw[draw=figaxis, line width=0.4pt]
  ([xshift=-144.31pt] axis cs:0,0.7) -- ([xshift=-140.31pt] axis cs:0,0.7);
\node[anchor=east, align=right, font=\figtick, text=figinksec, inner sep=0pt]
  at ([xshift=-151.71pt] axis cs:0,5.5) {Same draws\\select and score};
\node[anchor=east, align=right, font=\figtick, text=figinksec, inner sep=0pt]
  at ([xshift=-151.71pt] axis cs:0,2.5) {Held-out\\scoring};
\end{axis}
\node[anchor=south west, font=\figtitle, text=figink, inner sep=0pt]
  at (1.2pt,2.33in) {Held-out scoring results};
\path[use as bounding box]
  ([xshift=-1.1pt,yshift=-1.1pt]current bounding box.south west)
  rectangle
  ([xshift=1.1pt,yshift=1.1pt]current bounding box.north east);
\end{tikzpicture}%
\caption{\textbf{Held-out scoring.} Advantage of selected restart positions over
uniform placement when the same draws select and score (gray) and when selection
and scoring use disjoint draws (blue), with $95\%$ intervals where available.
The protocols estimate different selectors and cover different cells
(Appendix~\ref{sec:independent-audit}).}
\label{fig:ladder}
\end{figure}
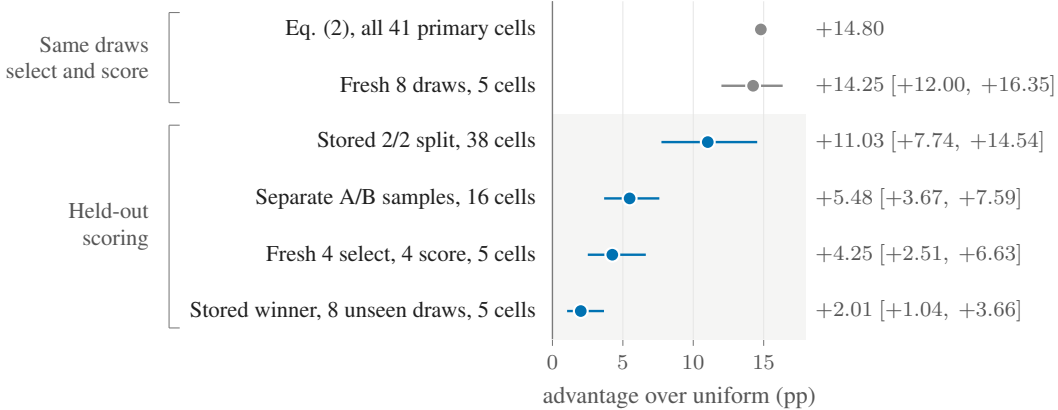

Figure~\ref{fig:ladder} scores selected positions on draws that did not select
them. The advantage shrinks, as selection on noise predicts, and persists. In a five-cell audit of $\FreshAuditExamples$ examples, positions
selected on four fresh continuations have a restart-success advantage over
uniform of $\FreshIndependentFour\pp$
$[\FreshIndependentFourLow,\FreshIndependentFourHigh]$ when scored on the
complementary four, and positions selected from the original four stored
continuations keep $\FreshStoredWinner\pp$
$[\FreshStoredWinnerLow,\FreshStoredWinnerHigh]$ on eight new continuations. Across
$\IndependentPrefixCells$ cells with two separately generated continuation
samples per prefix, the advantage is $\IndependentPrefixGapPP\pp$, positive in
every cell. Across the $\CrossfitAnalyzableCells$ primary cells that retain
individually labeled continuation outcomes, selecting on two of each step's
four stored continuations and scoring on the other two gives $\CrossfitGap\pp$
$[\CrossfitGapLow,\CrossfitGapHigh]$, positive in $\CrossfitPositiveCells$ of
$\CrossfitAnalyzableCells$ cells. These audits estimate different selectors
(Appendix~\ref{sec:independent-audit}, Table~\ref{tab:independent-scoring}),
but all place the selected positions above uniform placement in aggregate.

\SectionSpace
\section{Where to Restart Versus Whether to Spend}
\label{sec:policy-boundary}

We evaluate a learned router that, without observing restart outcomes, scores
each eligible step from text and structure features, process reward model (PRM) summaries, and token
log-probability summaries. It is trained with a listwise loss over the steps
of a chain and a learned depth prior, and it restricts candidates to a
\emph{benchmark-conditional late support}: the last third of the chain (the
middle third for Putnam). A separate confidence gate decides whether to spend a
restart on a chain at all.

We compare placement rules on held-out problems by gap recovered at a $10\%$
budget,
\begin{equation*}
  \text{GR@10}=\frac{\text{rule}-\text{uniform}}{\text{oracle}-\text{uniform}},
\end{equation*}
where each term is the mean observed restart rate at the positions a policy
selects. \emph{Uniform} selects eligible positions uniformly at random
(GR@10${}=0$), and \emph{oracle} is the same-draw top-$m_i$ selector of
Eq.~\eqref{eq:formal-gap} (GR@10${}=1$), so GR@10 is the fraction of the
oracle's advantage over random placement that a rule recovers. The simplest
fixed rule, \emph{always-last}, selects the last $m_i$ eligible positions.
GR@10 is computed on held-out problems whose restart rates vary across
positions; when the gate skips a chain, that chain contributes its base
outcome to every term.

\begin{figure}[H]
\centering
\begin{tikzpicture}
\path[use as bounding box] (0in,0in) rectangle (5.5in,2.32in);

\begin{axis}[
  premier,
  scale only axis,
  width=0.78in,height=1.42in,
  at={(1.83in,0.50in)},anchor=south west,
  axis x line*=bottom,
  axis y line=none,
  xmin=-0.04,xmax=0.16,
  ymin=0.35,ymax=5.75,
  xtick={0,0.10},
  xticklabel style={/pgf/number format/fixed,
                    /pgf/number format/fixed zerofill,
                    /pgf/number format/precision=2},
  ytick=\empty,
  ymajorgrids=false,
  xlabel={GR@10 difference},
  xlabel style={yshift=-1pt},
]
\draw[draw=none,fill=figgraytint]
  (axis cs:-0.04,2.75) rectangle (axis cs:0.16,5.55);
\draw[draw=figgrid,line width=0.4pt]
  (axis cs:0.10,0.35) -- (axis cs:0.10,5.75);
\draw[draw=figaxis,line width=0.5pt]
  (axis cs:0,0.35) -- (axis cs:0,5.75);
\draw[draw=figverm,line width=0.9pt]
  (axis cs:\RouterMinusUniformLow,5.15) -- (axis cs:\RouterMinusUniformHigh,5.15);
\draw[draw=figverm,line width=0.9pt]
  (axis cs:\RouterMinusLastUngatedLow,4.20) -- (axis cs:\RouterMinusLastUngatedHigh,4.20);
\draw[draw=figverm,line width=0.9pt]
  (axis cs:\RouterMinusLastGatedLow,3.25) -- (axis cs:\RouterMinusLastGatedHigh,3.25);
\draw[draw=figverm,line width=0.9pt]
  (axis cs:\GateEffectRouterLow,1.75) -- (axis cs:\GateEffectRouterHigh,1.75);
\draw[draw=figblue,line width=0.9pt]
  (axis cs:\GateEffectLastLow,0.80) -- (axis cs:\GateEffectLastHigh,0.80);
\addplot[only marks,mark=*,mark options={draw=white,line width=0.8pt,fill=figverm},
  mark size=2.5pt,forget plot] coordinates {
  (\RouterMinusUniform,5.15)
  (\RouterMinusLastUngated,4.20)
  (\RouterMinusLastGated,3.25)
  (\GateEffectRouter,1.75)};
\addplot[only marks,mark=*,mark options={draw=white,line width=0.8pt,fill=figblue},
  mark size=2.5pt,forget plot] coordinates {(\GateEffectLast,0.80)};
\node[anchor=east,inner sep=0pt,font=\figtick,text=figink]
  at ([xshift=-0.08in]axis cs:-0.04,5.15) {uniform};
\node[anchor=east,inner sep=0pt,font=\figtick,text=figink]
  at ([xshift=-0.08in]axis cs:-0.04,4.20) {always-last ungated};
\node[anchor=east,inner sep=0pt,font=\figtick,text=figink]
  at ([xshift=-0.08in]axis cs:-0.04,3.25) {always-last gated};
\node[anchor=east,inner sep=0pt,font=\figtick,text=figink]
  at ([xshift=-0.08in]axis cs:-0.04,1.75) {router};
\node[anchor=east,inner sep=0pt,font=\figtick,text=figink]
  at ([xshift=-0.08in]axis cs:-0.04,0.80) {always-last};
\draw[draw=figaxis,line width=0.4pt]
  ([xshift=-1.19in]axis cs:-0.04,5.45) -- ([xshift=-1.19in]axis cs:-0.04,2.95);
\draw[draw=figaxis,line width=0.4pt]
  ([xshift=-1.19in]axis cs:-0.04,5.45) -- ([xshift=-1.14in]axis cs:-0.04,5.45);
\draw[draw=figaxis,line width=0.4pt]
  ([xshift=-1.19in]axis cs:-0.04,2.95) -- ([xshift=-1.14in]axis cs:-0.04,2.95);
\draw[draw=figaxis,line width=0.4pt]
  ([xshift=-1.19in]axis cs:-0.04,2.05) -- ([xshift=-1.19in]axis cs:-0.04,0.55);
\draw[draw=figaxis,line width=0.4pt]
  ([xshift=-1.19in]axis cs:-0.04,2.05) -- ([xshift=-1.14in]axis cs:-0.04,2.05);
\draw[draw=figaxis,line width=0.4pt]
  ([xshift=-1.19in]axis cs:-0.04,0.55) -- ([xshift=-1.14in]axis cs:-0.04,0.55);
\node[anchor=east,align=right,inner sep=0pt,font=\figtick,text=figinksec]
  at ([xshift=-1.26in]axis cs:-0.04,4.20) {Router\\minus};
\node[anchor=east,align=right,inner sep=0pt,font=\figtick,text=figinksec]
  at ([xshift=-1.26in]axis cs:-0.04,1.30) {Gate effect\\on};
\end{axis}

\begin{axis}[
  premier,
  scale only axis,
  width=0.80in,height=1.42in,
  at={(4.15in,0.50in)},anchor=south west,
  axis x line*=bottom,
  axis y line=none,
  xmin=0,xmax=0.055,
  ymin=0.90,ymax=6.80,
  xtick={0,0.025,0.050},
  xticklabel style={/pgf/number format/fixed,
                    /pgf/number format/precision=3},
  ytick=\empty,
  ymajorgrids=false,
  xlabel={restart-success residual},
  xlabel style={yshift=-1pt},
]
\draw[draw=none,fill=figgraytint]
  (axis cs:0,1.08) rectangle (axis cs:0.055,2.95);
\draw[draw=figgrid,line width=0.4pt]
  (axis cs:0.025,1.05) -- (axis cs:0.025,6.65);
\draw[draw=figgrid,line width=0.4pt]
  (axis cs:0.050,1.05) -- (axis cs:0.050,6.65);
\draw[draw=figaxis,line width=0.5pt]
  (axis cs:0,1.05) -- (axis cs:0,6.65);
\draw[draw=figink,line width=0.9pt]
  (axis cs:\DepthResidualALow,6.34) -- (axis cs:\DepthResidualAHigh,6.34);
\draw[draw=figink,line width=0.9pt]
  (axis cs:\DepthResidualBLow,5.28) -- (axis cs:\DepthResidualBHigh,5.28);
\draw[draw=figink,line width=0.9pt]
  (axis cs:\RichResidualAProblemLow,2.56) -- (axis cs:\RichResidualAProblemHigh,2.56);
\draw[draw=figink,line width=0.9pt]
  (axis cs:\RichResidualBProblemLow,1.50) -- (axis cs:\RichResidualBProblemHigh,1.50);
\addplot[only marks,mark=*,mark options={draw=white,line width=0.8pt,fill=figink},
  mark size=2.5pt,forget plot] coordinates {
  (\DepthResidualA,6.34) (\DepthResidualB,5.28)
  (\RichResidualA,2.56) (\RichResidualB,1.50)};
\node[anchor=east,inner sep=0pt,font=\figtick,text=figink]
  at ([xshift=-0.08in]axis cs:0,6.34) {A to B};
\node[anchor=east,inner sep=0pt,font=\figtick,text=figink]
  at ([xshift=-0.08in]axis cs:0,5.28) {B to A};
\node[anchor=east,inner sep=0pt,font=\figtick,text=figink]
  at ([xshift=-0.08in]axis cs:0,2.56) {A to B};
\node[anchor=east,inner sep=0pt,font=\figtick,text=figink]
  at ([xshift=-0.08in]axis cs:0,1.50) {B to A};
\node[anchor=east,inner sep=0pt,font=\figtick,text=figinksec]
  at (axis cs:0.084,6.34) {$\DepthResidualAPositiveCells/\DepthResidualCells$};
\node[anchor=east,inner sep=0pt,font=\figtick,text=figinksec]
  at (axis cs:0.084,5.28) {$\DepthResidualBPositiveCells/\DepthResidualCells$};
\node[anchor=east,inner sep=0pt,font=\figtick,text=figinksec]
  at (axis cs:0.084,2.56) {$\RichResidualAPositiveCells/\DepthResidualCells$};
\node[anchor=east,inner sep=0pt,font=\figtick,text=figinksec]
  at (axis cs:0.084,1.50) {$\RichResidualBPositiveCells/\DepthResidualCells$};
\draw[draw=figaxis,line width=0.4pt]
  ([xshift=-0.65in]axis cs:0,6.60) -- ([xshift=-0.65in]axis cs:0,4.85);
\draw[draw=figaxis,line width=0.4pt]
  ([xshift=-0.65in]axis cs:0,6.60) -- ([xshift=-0.60in]axis cs:0,6.60);
\draw[draw=figaxis,line width=0.4pt]
  ([xshift=-0.65in]axis cs:0,4.85) -- ([xshift=-0.60in]axis cs:0,4.85);
\draw[draw=figaxis,line width=0.4pt]
  ([xshift=-0.65in]axis cs:0,2.95) -- ([xshift=-0.65in]axis cs:0,1.10);
\draw[draw=figaxis,line width=0.4pt]
  ([xshift=-0.65in]axis cs:0,2.95) -- ([xshift=-0.60in]axis cs:0,2.95);
\draw[draw=figaxis,line width=0.4pt]
  ([xshift=-0.65in]axis cs:0,1.10) -- ([xshift=-0.60in]axis cs:0,1.10);
\node[anchor=east,align=right,inner sep=0pt,font=\figtick,text=figinksec]
  at ([xshift=-0.76in]axis cs:0,5.725) {Depth class};
\node[anchor=east,align=right,inner sep=0pt,font=\figtick,text=figinksec]
  at ([xshift=-0.76in]axis cs:0,2.025) {Enriched\\class};
\end{axis}

\node[anchor=south west,inner sep=0pt,font=\figtitle,text=figink]
  at (0.05in,2.10in) {(a) Placement and spending};
\node[anchor=south west,inner sep=0pt,font=\figtitle,text=figink]
  at (2.82in,2.10in) {(b) Oracle headroom};
\end{tikzpicture}%
\caption{\textbf{Placement, spending, and oracle headroom.} (a) Paired GR@10
differences over $\RouterUniformSeeds$ split seeds: the router against uniform
placement (both ungated) and against always-last (both ungated, and both
gated), and the gate's effect on each policy (orange involves the router, blue is
always-last). Intervals are split-seed
stability intervals on one frozen store, not problem-population intervals.
(b) Held-out restart-success residual of cross-fitted oracle selection beyond
the declared 21-policy depth class and the enriched positional class, in both
directions (A to B: sample A selects, sample B scores); numbers at right count
cells with a positive residual out of $\DepthResidualCells$. Intervals
resample examples pooled across cells (depth class) or problem clusters
(enriched class). This is
an oracle target, not a feasible-policy gain. Exact values are in Appendix
Tables~\ref{tab:decisions}, \ref{tab:depth-residual},
and~\ref{tab:rich-residual}.}
\label{fig:decisions}
\end{figure}
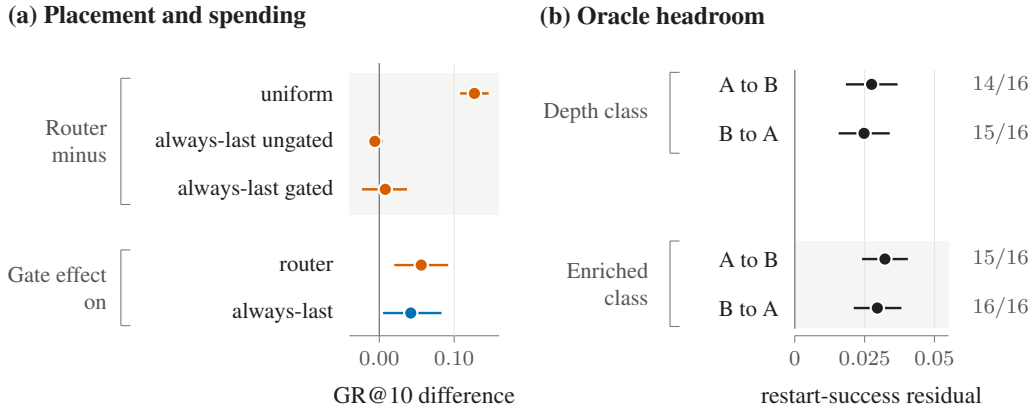

Figure~\ref{fig:decisions}a separates where to restart from whether to spend
over 25 split seeds. Ungated, the router beats uniform placement by
$\RouterMinusUniform$ GR@10 $[\RouterMinusUniformLow,\RouterMinusUniformHigh]$,
in $\RouterUniformWins$ of $\RouterUniformSeeds$ seeds. Against always-last,
however, it shows no detected gain: the paired difference is
$\RouterMinusLastUngated$
$[\RouterMinusLastUngatedLow,\RouterMinusLastUngatedHigh]$ ungated and
$\RouterMinusLastGated$
$[\RouterMinusLastGatedLow,\RouterMinusLastGatedHigh]$ gated. Gate-on minus gate-off
GR@10 is $\GateEffectRouter$ $[\GateEffectRouterLow,\GateEffectRouterHigh]$ for
the router and $\GateEffectLast$ $[\GateEffectLastLow,\GateEffectLastHigh]$ for
always-last, and the paired difference between the two is $\GateDifference$
$[\GateDifferenceLow,\GateDifferenceHigh]$. A separate fixed-composition audit, comparing learned
gating with random chain selection under always-last, finds an unresolved
accuracy contrast (Appendix Table~\ref{tab:decisions}). The router's gain over uniform is therefore not separable from late position. Component controls agree: dropping the late support entirely (while
keeping the learned depth prior) changes little, removing the depth prior
improves GR@10 by $\PriorRemoval$, and adding PRM, log-probability, or
hidden-state features shows no detected improvement in the tested selector
within the evaluated scopes (Appendix Table~\ref{tab:requested-components}).

Always-last is therefore the reference that learned placement must beat:
the router beats uniform placement in every seed yet shows no detected gain over
always-last. Gates must also be applied to every compared policy, because the
gate raised GR@10 for both policies ($\GateEffectRouter$ and $\GateEffectLast$);
comparing a gated learned policy with an ungated baseline would credit spending
to placement.

\SectionSpace
\section{Oracle Headroom Beyond Declared Positional Classes}
\label{sec:residual}

Cross-fitted oracle selection measures the held-out restart-success residual
over the rule selected from each declared positional class; the residual is an
oracle target, not a feasible selector's gain. We use the two
continuation samples collected at each prefix, A and B, from the point- and
block-restart sweeps (Appendices~\ref{sec:independent-audit}
and~\ref{sec:policy-protocol}). Sample A fits a declared positional class and
selects the oracle's restart positions; sample B scores both, and the roles are
then swapped. On the restart positions $1$ to $K_i-3$ that the two sweeps share, across
$\DepthResidualCells$ cells, the oracle exceeds the rule selected from a
21-policy depth class by $\DepthResidualA$
$[\DepthResidualALow,\DepthResidualAHigh]$ and $\DepthResidualB$
$[\DepthResidualBLow,\DepthResidualBHigh]$ in held-out restart-success rate,
positive in $\DepthResidualAPositiveCells$ and $\DepthResidualBPositiveCells$
of $\DepthResidualCells$ cells (Figure~\ref{fig:decisions}b; Appendix
Table~\ref{tab:depth-residual}). The residual remains positive against a richer
positional learner, which scores relative position separately by chain length
(length-by-rank tables) with model selection nested inside the training folds;
that learner does not outperform the original class out of sample
(Table~\ref{tab:rich-residual}).

Because continuation outcomes select the oracle's positions, the residual is an
oracle target for feasible selectors, lying beyond the declared positional
classes (the richer family does not span every positional function). Whether
features can predict it is open: the PRM, log-probability, and hidden-state
summaries we added showed no detected improvement in the tested selector
(Table~\ref{tab:requested-components}).

\SectionSpace
\section{Always-Last Versus the Self-Consistency Frontier}
\label{sec:frontier}

We compare always-last with the standard way of spending extra samples,
self-consistency (SC), which samples $N$ complete chains and returns the most
common parsed answer. We compare against the exact SC
frontier. At each policy's aggregate generated output, we solve for the best
randomized mixture of SC(1) to SC(4) with the same output, the best comparator
in the tested SC(1) to SC(4) family at that budget. The panel is a frozen
DeepSeek-R1-Distill-Qwen-14B/MATH-500 set of $\FrontierProblems$ problems and
$\FrontierSeeds$ generation seeds (Appendix~\ref{sec:frontier-protocol}).

\begin{figure}[H]
\centering
\begin{tikzpicture}[x=1in,y=1in]
\path[use as bounding box] (0,0) rectangle (5.5,1.88);

\def\FfigAccLeft{1.08}
\def\FfigAccRight{3.42}
\def\FfigOutLeft{3.92}
\def\FfigOutRight{5.34}
\def\FfigAccBottom{0.45}
\def\FfigPlotTop{1.58}
\def\FfigSCY{1.42}
\def\FfigAlwaysLastY{1.00}
\def\FfigRouterY{0.58}

\pgfmathsetmacro{\FfigSCFourAccX}{\FfigAccLeft + (\SCKFourAccuracy - 0.76) * (\FfigAccRight - \FfigAccLeft) / 0.09}
\pgfmathsetmacro{\FfigAlwaysLastFrontierX}{\FfigAccLeft + (\AlwaysLastFrontierAccuracy - 0.76) * (\FfigAccRight - \FfigAccLeft) / 0.09}
\pgfmathsetmacro{\FfigAlwaysLastAccX}{\FfigAccLeft + (\AlwaysLastAccuracy - 0.76) * (\FfigAccRight - \FfigAccLeft) / 0.09}
\pgfmathsetmacro{\FfigRouterFrontierX}{\FfigAccLeft + (\RouterFrontierAccuracy - 0.76) * (\FfigAccRight - \FfigAccLeft) / 0.09}
\pgfmathsetmacro{\FfigRouterAccX}{\FfigAccLeft + (\RouterAccuracy - 0.76) * (\FfigAccRight - \FfigAccLeft) / 0.09}
\pgfmathsetmacro{\FfigSCFourOutX}{\FfigOutLeft + \SCKFourRatio * (\FfigOutRight - \FfigOutLeft) / 1.05}
\pgfmathsetmacro{\FfigAlwaysLastOutX}{\FfigOutLeft + \AlwaysLastOutputRatio * (\FfigOutRight - \FfigOutLeft) / 1.05}
\pgfmathsetmacro{\FfigRouterOutX}{\FfigOutLeft + \RouterOutputRatio * (\FfigOutRight - \FfigOutLeft) / 1.05}

\node[anchor=south west,inner sep=0pt,font=\figtitle,text=figink]
  at (\FfigAccLeft,1.66) {Accuracy};
\node[anchor=south west,inner sep=0pt,font=\figtitle,text=figink]
  at (\FfigOutLeft,1.66) {Generated output};

\foreach \tick/\label in {0.76/0.76,0.78/0.78,0.80/0.80,0.82/0.82,0.84/0.84}{%
  \pgfmathsetmacro{\FfigTickX}{\FfigAccLeft + (\tick - 0.76) * (\FfigAccRight - \FfigAccLeft) / 0.09}%
  \draw[draw=figgrid,line width=0.4pt] (\FfigTickX,\FfigAccBottom) -- (\FfigTickX,\FfigPlotTop);%
  \draw[draw=figaxis,line width=0.4pt] (\FfigTickX,\FfigAccBottom) -- (\FfigTickX,0.42);%
  \node[anchor=north,inner sep=0pt,font=\figtick,text=figinksec]
    at (\FfigTickX,0.35) {\label};%
}
\draw[draw=figaxis,line width=0.45pt]
  (\FfigAccLeft,\FfigAccBottom) -- (\FfigAccRight,\FfigAccBottom);

\foreach \tick/\label in {0/0,0.5/0.5,1.0/1.0}{%
  \pgfmathsetmacro{\FfigOutTickX}{\FfigOutLeft + \tick * (\FfigOutRight - \FfigOutLeft) / 1.05}%
  \ifdim \tick pt=0pt\relax%
    \draw[draw=figaxis,line width=0.45pt] (\FfigOutTickX,\FfigAccBottom) -- (\FfigOutTickX,0.42);%
  \else%
    \draw[draw=figgrid,line width=0.4pt] (\FfigOutTickX,\FfigAccBottom) -- (\FfigOutTickX,\FfigPlotTop);%
    \draw[draw=figaxis,line width=0.4pt] (\FfigOutTickX,\FfigAccBottom) -- (\FfigOutTickX,0.42);%
  \fi%
  \node[anchor=north,inner sep=0pt,font=\figtick,text=figinksec]
    at (\FfigOutTickX,0.35) {\label};%
}
\draw[draw=figaxis,line width=0.45pt]
  (\FfigOutLeft,\FfigAccBottom) -- (\FfigOutRight,\FfigAccBottom);
\node[anchor=south,inner sep=0pt,font=\figlabel,text=figinksec]
  at ({(\FfigOutLeft+\FfigOutRight)/2},0.055) {generated output (\(\times\) SC(4))};

\node[anchor=east,inner sep=0pt,font=\figtick,text=figink]
  at (0.90,\FfigSCY) {SC(4)};
\node[anchor=east,inner sep=0pt,font=\figtick,text=figink]
  at (0.90,\FfigAlwaysLastY) {always-last};
\node[anchor=east,inner sep=0pt,font=\figtick,text=figink]
  at (0.90,\FfigRouterY) {router};

\draw[draw=figbluesoft,line width=2.7pt,line cap=round]
  (\FfigAlwaysLastFrontierX,\FfigAlwaysLastY) -- (\FfigAlwaysLastAccX,\FfigAlwaysLastY);
\draw[draw=figvermsoft,line width=2.7pt,line cap=round]
  (\FfigRouterFrontierX,\FfigRouterY) -- (\FfigRouterAccX,\FfigRouterY);
\node[circle,draw=figmuted,fill=white,line width=0.8pt,inner sep=0pt,minimum size=5.6pt]
  at (\FfigAlwaysLastFrontierX,\FfigAlwaysLastY) {};
\node[circle,draw=figmuted,fill=white,line width=0.8pt,inner sep=0pt,minimum size=5.6pt]
  at (\FfigRouterFrontierX,\FfigRouterY) {};
\node[circle,draw=white,fill=figmuted,line width=0.8pt,inner sep=0pt,minimum size=6pt]
  at (\FfigSCFourAccX,\FfigSCY) {};
\node[circle,draw=white,fill=figblue,line width=0.8pt,inner sep=0pt,minimum size=6pt]
  at (\FfigAlwaysLastAccX,\FfigAlwaysLastY) {};
\node[circle,draw=white,fill=figverm,line width=0.8pt,inner sep=0pt,minimum size=6pt]
  at (\FfigRouterAccX,\FfigRouterY) {};

\draw[draw=figmuted,line width=0.4pt]
  (\FfigAlwaysLastFrontierX,1.04) -- (\FfigAlwaysLastFrontierX,1.14);
\node[anchor=south,fill=white,inner sep=2pt,font=\figtick,text=figinksec]
  at (\FfigAlwaysLastFrontierX,1.19) {SC at same output};
\node[anchor=south east,inner sep=0pt,font=\figtick,text=figblue]
  at (\FfigAlwaysLastAccX,1.12) {\AlwaysLastFrontierDelta};
\node[anchor=south east,inner sep=0pt,font=\figtick,text=figverm]
  at (\FfigRouterAccX,0.70) {\RouterFrontierDelta};

\draw[draw=figmuted,line width=1.25pt,line cap=round]
  (\FfigOutLeft,\FfigSCY) -- (\FfigSCFourOutX,\FfigSCY);
\draw[draw=figblue,line width=1.25pt,line cap=round]
  (\FfigOutLeft,\FfigAlwaysLastY) -- (\FfigAlwaysLastOutX,\FfigAlwaysLastY);
\draw[draw=figverm,line width=1.25pt,line cap=round]
  (\FfigOutLeft,\FfigRouterY) -- (\FfigRouterOutX,\FfigRouterY);
\node[circle,draw=white,fill=figmuted,line width=0.6pt,inner sep=0pt,minimum size=4.2pt]
  at (\FfigSCFourOutX,\FfigSCY) {};
\node[circle,draw=white,fill=figblue,line width=0.6pt,inner sep=0pt,minimum size=4.2pt]
  at (\FfigAlwaysLastOutX,\FfigAlwaysLastY) {};
\node[circle,draw=white,fill=figverm,line width=0.6pt,inner sep=0pt,minimum size=4.2pt]
  at (\FfigRouterOutX,\FfigRouterY) {};
\node[anchor=south east,inner sep=0pt,font=\figtick,text=figmuted]
  at (\FfigSCFourOutX,1.52) {\SCKFourRatio\(\times\)};
\node[anchor=south east,inner sep=0pt,font=\figtick,text=figblue]
  at (\FfigAlwaysLastOutX,1.10) {\AlwaysLastOutputRatio\(\times\)};
\node[anchor=south east,inner sep=0pt,font=\figtick,text=figverm]
  at (\FfigRouterOutX,0.68) {\RouterOutputRatio\(\times\)};
\end{tikzpicture}%
\caption{\textbf{Always-last is more accurate than self-consistency at matched
aggregate generated output in the tested cell} (DeepSeek-R1-Distill-Qwen-14B/MATH-500,
$\FrontierProblems$ problems, $\FrontierSeeds$ seeds). Left: hollow circles are
the best mix of SC(1) to SC(4) at each policy's own generated output; filled
circles are the policy. Margins are $\AlwaysLastFrontierDelta$
$[\AlwaysLastFrontierDeltaLow,\AlwaysLastFrontierDeltaHigh]$ for always-last and
$\RouterFrontierDelta$ $[\RouterFrontierDeltaLow,\RouterFrontierDeltaHigh]$ for
the router ($95\%$ paired bootstrap; Appendix Table~\ref{tab:frontier}); their
difference is not detected. Right: aggregate generated output relative to
SC(4), not total tokens or wall-clock time.}
\label{fig:frontier}
\end{figure}
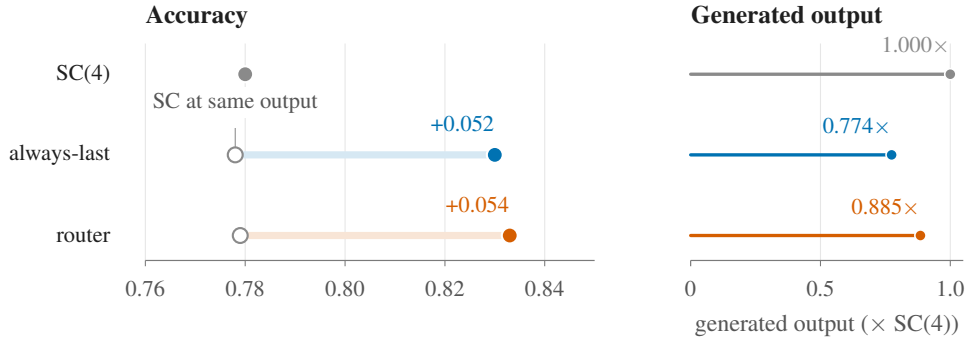

Always-last reaches $\AlwaysLastAccuracy$ accuracy at
$\AlwaysLastOutputRatio\times$ the aggregate generated output of SC(4) and
exceeds the exact SC(1) to SC(4) frontier at matched aggregate output by
$\AlwaysLastFrontierDelta$
$[\AlwaysLastFrontierDeltaLow,\AlwaysLastFrontierDeltaHigh]$; it lies above
the frontier in $\AlwaysLastSeedsAboveFrontier$ of
$\FrontierSeeds$ generation seeds (Figure~\ref{fig:frontier}; Appendix
Table~\ref{tab:frontier}). The router also clears its frontier, in
$\RouterSeedsAboveFrontier$ of $\FrontierSeeds$ seeds, and the paired
router-minus-always-last difference is unresolved, so the experiment supports
late continuation rather than learned restart placement, in one cell and under
one cost ledger. In a separate two-seed, 100-problem H200 assay, always-last used
$\HtwoTotalTokenRatio\times$ the aggregate input-plus-output tokens and
$\HtwoWallclockRatio\times$ the wall-clock time of parallel SC, with unresolved
accuracy; these results do not establish a latency or total-token benefit.

Always-last requires no training, verifier, or features, and a late restart
regenerates only the tail of an existing chain, which matters when generated
output is the binding cost. It runs after the base chain, so latency and total
tokens must be measured separately, and in the one other available cell,
Qwen3-32B/MATH-500, its margin is unresolved.

\SectionSpace
\section{Label Ties Can Manufacture Position Effects}
\label{sec:loss}

Our pointwise step selector is trained on binary labels: with four rollouts
per step, the top $m_i$ steps of each chain are labeled positive. Several steps
often tie at the selection cutoff, and breaking those ties by earliest index
favors earlier steps, injecting an artificial early-position signal into the
labels (Figure~\ref{fig:tie-correction}a). In a separate five-seed diagnostic
without positional support, mean pointwise binary cross-entropy (BCE) GR@10
moves from $\TieBCEFirst$ under earliest-index ties to $\TieBCERandom$ under
randomized ties, and the sign flips in all $\TieSeeds$ seeds (ranges
$\TieBCEFirstLow$ to $\TieBCEFirstHigh$ and $\TieBCERandomLow$ to
$\TieBCERandomHigh$). The evaluated ListNet formulation~\citep{cao2007listnet}, which trains
on within-chain relative scores, is exactly invariant ($\TieListNetFirst$ under
both rules; Figure~\ref{fig:tie-correction}b).

\begin{figure}[H]
\centering
\begin{tikzpicture}
\path[use as bounding box] (0,0) rectangle (5.5in,2.04in);

\node[anchor=south west,inner sep=0pt,font=\figtitle,text=figink]
  at (0.05in,1.90in) {(a) Tie rule};
\node[anchor=east,inner sep=0pt,font=\figtick,text=figink]
  at (0.79in,1.54in) {correct / 4};
\node[anchor=east,inner sep=0pt,font=\figtick,text=figink]
  at (0.79in,1.00in) {earliest index};
\node[anchor=east,inner sep=0pt,font=\figtick,text=figink]
  at (0.79in,0.60in) {randomized};

\path[draw=figaxis,line width=0.4pt,fill=figink!12!white]
  (0.86in,1.42in) rectangle (1.02in,1.66in);
\node[font=\figtick,text=figink,inner sep=0pt] at (0.94in,1.54in) {1};
\path[draw=figink,line width=0.8pt,fill=figink!28!white]
  (1.055in,1.42in) rectangle (1.215in,1.66in);
\node[font=\figtick,text=figink,inner sep=0pt] at (1.135in,1.54in) {4};
\path[draw=figaxis,line width=0.4pt,fill=figgraytint]
  (1.25in,1.42in) rectangle (1.41in,1.66in);
\node[font=\figtick,text=figink,inner sep=0pt] at (1.33in,1.54in) {0};
\path[draw=figaxis,line width=0.4pt,fill=figink!18!white]
  (1.445in,1.42in) rectangle (1.605in,1.66in);
\node[font=\figtick,text=figink,inner sep=0pt] at (1.525in,1.54in) {2};
\path[draw=figink,line width=0.8pt,fill=figink!28!white]
  (1.64in,1.42in) rectangle (1.80in,1.66in);
\node[font=\figtick,text=figink,inner sep=0pt] at (1.72in,1.54in) {4};
\path[draw=figaxis,line width=0.4pt,fill=figink!12!white]
  (1.835in,1.42in) rectangle (1.995in,1.66in);
\node[font=\figtick,text=figink,inner sep=0pt] at (1.915in,1.54in) {1};
\path[draw=figaxis,line width=0.4pt,fill=figink!22!white]
  (2.03in,1.42in) rectangle (2.19in,1.66in);
\node[font=\figtick,text=figink,inner sep=0pt] at (2.11in,1.54in) {3};
\path[draw=figink,line width=0.8pt,fill=figink!28!white]
  (2.225in,1.42in) rectangle (2.385in,1.66in);
\node[font=\figtick,text=figink,inner sep=0pt] at (2.305in,1.54in) {4};

\draw[draw=white,fill=figink,line width=0.8pt]
  (1.135in,1.00in) circle[radius=2.7pt];
\draw[draw=figink,fill=white,line width=0.8pt]
  (1.135in,0.60in) circle[radius=2.9pt];
\draw[draw=figink,fill=white,line width=0.8pt]
  (1.72in,0.60in) circle[radius=2.9pt];
\draw[draw=figink,fill=white,line width=0.8pt]
  (2.305in,0.60in) circle[radius=2.9pt];
\draw[-{Latex[length=3pt,width=2pt]},draw=figaxis,line width=0.45pt]
  (0.94in,0.27in) -- (2.305in,0.27in);
\node[anchor=north west,inner sep=0pt,font=\figtick,text=figinksec]
  at (0.94in,0.22in) {early};
\node[anchor=north east,inner sep=0pt,font=\figtick,text=figinksec]
  at (2.305in,0.22in) {late};

\node[anchor=south west,inner sep=0pt,font=\figtitle,text=figink]
  at (2.67in,1.90in) {(b) GR@10};
\begin{axis}[
  premier,
  scale only axis,
  at={(4.00in,0.40in)},anchor=south west,
  width=1.22in,height=1.12in,
  xmin=-0.25,xmax=0.22,
  ymin=0.35,ymax=3.85,
  axis x line*=bottom,
  axis y line=none,
  xtick={-0.20,0,0.20},
  xticklabel style={/pgf/number format/fixed,
                    /pgf/number format/precision=2},
  ytick=\empty,
  xmajorgrids=false,ymajorgrids=false,
  xlabel={GR@10},
  xlabel style={yshift=1pt},
]
\draw[draw=figgrid,line width=0.4pt]
  (axis cs:-0.10,0.42) -- (axis cs:-0.10,3.55);
\draw[draw=figgrid,line width=0.4pt]
  (axis cs:0.10,0.42) -- (axis cs:0.10,1.08);
\draw[draw=figgrid,line width=0.4pt]
  (axis cs:0.10,2.93) -- (axis cs:0.10,3.55);
\draw[draw=figaxis,line width=0.55pt]
  (axis cs:0,0.42) -- (axis cs:0,3.55);
\node[anchor=south,inner sep=0pt,font=\figtick,text=figinksec]
  at (axis cs:0,3.67) {uniform};

\path[draw=figink,line width=0.8pt,fill=white]
  (axis cs:\TieBCEFirst,2.83) rectangle (axis cs:0,3.27);
\path[draw=figmuted,line width=0.55pt,fill=figmuted!22!white]
  (axis cs:0,1.73) rectangle (axis cs:\TieBCERandom,2.17);
\path[draw=figmuted,line width=0.55pt,fill=figmuted!22!white]
  (axis cs:0,0.63) rectangle (axis cs:\TieListNetFirst,1.07);

\draw[draw=figink,line width=0.75pt]
  (axis cs:\TieBCEFirstLow,3.05) -- (axis cs:\TieBCEFirstHigh,3.05);
\draw[draw=figink,line width=0.75pt]
  (axis cs:\TieBCEFirstLow,2.93) -- (axis cs:\TieBCEFirstLow,3.17);
\draw[draw=figink,line width=0.75pt]
  (axis cs:\TieBCEFirstHigh,2.93) -- (axis cs:\TieBCEFirstHigh,3.17);
\draw[draw=figinksec,line width=0.75pt]
  (axis cs:\TieBCERandomLow,1.95) -- (axis cs:\TieBCERandomHigh,1.95);
\draw[draw=figinksec,line width=0.75pt]
  (axis cs:\TieBCERandomLow,1.83) -- (axis cs:\TieBCERandomLow,2.07);
\draw[draw=figinksec,line width=0.75pt]
  (axis cs:\TieBCERandomHigh,1.83) -- (axis cs:\TieBCERandomHigh,2.07);
\draw[draw=figinksec,line width=0.75pt]
  (axis cs:\TieListNetLow,0.85) -- (axis cs:\TieListNetHigh,0.85);
\draw[draw=figinksec,line width=0.75pt]
  (axis cs:\TieListNetLow,0.73) -- (axis cs:\TieListNetLow,0.97);
\draw[draw=figinksec,line width=0.75pt]
  (axis cs:\TieListNetHigh,0.73) -- (axis cs:\TieListNetHigh,0.97);

\node[anchor=east,inner sep=0pt,font=\figtick,text=figink]
  at ([xshift=-7pt]axis cs:-0.25,3.05) {BCE, earliest index};
\node[anchor=east,inner sep=0pt,font=\figtick,text=figink]
  at ([xshift=-7pt]axis cs:-0.25,1.95) {BCE, randomized};
\node[anchor=east,inner sep=0pt,font=\figtick,text=figink]
  at ([xshift=-7pt]axis cs:-0.25,0.85) {ListNet, either rule};
\node[anchor=south,inner sep=0pt,font=\figtick,text=figink]
  at (axis cs:\TieBCEFirst,3.46) {$\TieBCEFirst$};
\node[anchor=south,inner sep=0pt,font=\figtick,text=figink]
  at (axis cs:\TieBCERandom,2.36) {$\TieBCERandom$};
\node[anchor=south,inner sep=0pt,font=\figtick,text=figink]
  at (axis cs:\TieListNetFirst,1.26) {$\TieListNetFirst$};
\end{axis}
\end{tikzpicture}%
\caption{\textbf{Label ties can flip a pointwise selector.} (a) Schematic: one
chain's correct continuations out of four at each step (darker cells are
higher; outlined cells tie at the maximum). Earliest-index tie breaking always labels the first tied step,
which injects an early-position signal; randomized tie breaking labels one tied
step at random. (b) GR@10 in a five-seed diagnostic (bars are means; whiskers span the
minimum to maximum across seeds), where zero is uniform placement:
pointwise BCE moves
from $\TieBCEFirst$ with earliest-index ties to $\TieBCERandom$ with randomized
ties, while ListNet is $\TieListNetFirst$ under either rule. Fair-label loss
contrasts are in Appendix Table~\ref{tab:loss-extension}.}
\label{fig:tie-correction}
\end{figure}

In the separate primary 24-seed comparison with randomized ties, ListNet minus
BCE is $\FairLossUnmasked$ $[\FairLossUnmaskedLow,\FairLossUnmaskedHigh]$
without positional support and $\FairLossInBand$
$[\FairLossInBandLow,\FairLossInBandHigh]$ within the benchmark-conditional
late support, using a problem-cluster bootstrap (Appendix
Table~\ref{tab:loss-extension}). Both intervals cross zero, so we conclude
only that ListNet is invariant to tie resolution, not that one loss ranks above
the other or that they are equivalent within the support. After the primary results were
available, we locked the code, data, and configuration and ran a 64-seed
robustness extension. It meets its frozen criteria for a positive contrast
without positional support and for equivalence within
$\pm\FairLossEquivalenceMargin$ GR@10 inside the late support; it is a
post-lock robustness extension, not independent confirmation.

This pitfall and the attenuation in Section~\ref{sec:measurement} both bear on
process supervision, where steps are often labeled with Monte Carlo rollouts
from prefixes~\citep{math_shepherd,luo2024improve}. On the same five audited cells, selecting and
scoring on the same eight continuations gives $\FreshSameDraw\pp$, whereas
selecting on four and scoring on the complementary four gives
$\FreshIndependentFour\pp$. The selectors differ, but the attenuation is in the
direction predicted by selection on noise, so rollout-derived step importance
should be validated on held-out rollouts before it is used for training or
reported. Because ties are common with four rollouts per step, randomized tie
breaking removes systematic earliest-index labeling and the evaluated ListNet
loss is invariant to tie resolution; neither requires additional rollouts.

\SectionSpace
\section{Related Work}
\label{sec:related}

Self-consistency allocates compute across independent
chains~\citep{wang2022selfconsistency}, and inference-scaling studies compare
search and verification strategies under
budgets~\citep{snell2024scaling,brown2024large}. Process supervision and PRMs
score intermediate reasoning~\citep{lightman2023lets,uesato2022solving}, while
tree-based methods branch over partial
trajectories~\citep{yao2023tree,luo2024improve}. Resampling from intermediate
prefixes has also been used to locate forking tokens and influential reasoning
sentences~\citep{bigelow2025forking,bogdan2025thought}, and adaptive computation
methods learn when to halt or route
computation~\citep{graves2016adaptive,schuster2022confident}. We instead
measure interventional restart effects at every stored prefix, then test how
much of that opportunity fixed placement, declared positional classes, and a
feasible selector capture.

\SectionSpace
\section{Discussion}
\label{sec:discussion}

\paragraph{Where to continue is its own allocation decision}
Test-time scaling is usually framed as how many independent chains to sample
or how to rank complete candidates. Our measurements quantify a complementary
axis: where inside an existing chain to spend a continuation. Restart value
along this axis is uneven within chains, and its advantage survives held-out
scoring across models and benchmarks at reduced magnitude
(Section~\ref{sec:measurement}). Where to continue is therefore a
separate allocation decision: always-last is the reference baseline, and
same-draw optimism and tie-induced position bias are its characteristic
failure modes.

\paragraph{Always-last as the reference}
The learned router we evaluated shows no detected gain over a rule with no
learned component (Section~\ref{sec:policy-boundary}), and that rule clears
matched
self-consistency in the one tested cell under aggregate generated output
(Section~\ref{sec:frontier}). Claims of learned within-chain placement should
be measured against it. We recommend that such methods state whether a
reported gap is same-draw, held-out, or achieved by a feasible selector;
report paired contrasts against both uniform placement and always-last, with
any gate applied to every policy; and report generated output, total tokens,
and wall-clock time separately.

\paragraph{Restart value as a supervision target}
Restart value is an interventional quantity rather than a judgment of whether a
step is correct, and adding Math-Shepherd PRM summaries showed no detected
improvement in the tested selector. Restart value may therefore be a
useful supervision target, and the held-out residual beyond declared
positional classes (Section~\ref{sec:residual}) is the reference against which
a selector trained on it should be measured.

\SectionSpace
\section{Limitations and Broader Impact}
\label{sec:limitations}

Eq.~\eqref{eq:formal-gap} is a same-draw statistic, and the held-out audits
that check it cover fewer cells and estimate different selectors. Results are
conditional on a designed rather than sampled grid of models and benchmarks,
on automatic graders, and on stored step boundaries. The resolved self-consistency advantage is on one panel and is measured in
aggregate generated output; total tokens and wall-clock time are reported
separately, and the residual of Section~\ref{sec:residual} is an oracle target.
On the tested panel, continuing from stored prefixes was more accurate than the
self-consistency frontier at lower generated output; a miscalibrated selector
could instead damage correct answers or waste compute. The work uses public benchmarks and publicly released models and
collects no human-subject data.

\SectionSpace
\section{Conclusion}
\label{sec:conclusion}

Restart maps expose a second axis of test-time allocation: where to continue
an existing chain. Across 41 model and benchmark cells, restart value is uneven
within chains, and held-out audits on 5, 16, and 38 cells each retain a
positive advantage over uniform placement. Always-last is a strong fixed
baseline: our learned router shows no detected gain over it, and on
DeepSeek-R1-Distill-Qwen-14B/MATH-500 it exceeds the exact self-consistency
frontier by $\AlwaysLastFrontierDelta$ at matched aggregate generated output.
Cross-fitted oracle analysis shows that the declared positional classes do not
exhaust restart value, and randomized ties remove an early-position labeling
bias without additional rollouts. Future restart selectors should report
held-out utility, paired comparisons with always-last under matched gating,
and costs in explicit units.

\section*{Acknowledgments, Funding, and Competing Interests}
The authors thank the anonymous reviewers and area chairs for the controls that
strengthened this analysis. \textbf{Funding:} This work was self-funded
by the authors and received no external funding. \textbf{Competing interests:}
The authors declare no competing interests.

\bibliographystyle{plainnat}
\bibliography{references}

\appendix
\setcounter{topnumber}{2}
\setcounter{bottomnumber}{1}
\setcounter{totalnumber}{3}

\FloatBarrier
\section{Measurement Definitions and Cell Inclusion}
\label{sec:correction-record}

Eq.~\eqref{eq:formal-gap} is the budgeted eligible-position advantage
($\EqThreeGapPP\pp$). The maximum-observed-utility statistic, which averages
the largest observed utility per example, is a distinct quantity
($\MaximumObservedUtilityPP\pp$). All main-text tables and
figures are generated from checksummed JSON by a single build script
(Appendix~\ref{sec:reproducibility}).

A chain enters a cell when it has at least two eligible restart positions.
A cell is degenerate when its chains are effectively single-step
(Llama-Nemotron-70B on HMMT) or the model emits no usable multistep traces
(Nemotron-Nano-8B on GPQA); of the 54 model and benchmark pairs, these two are
degenerate and 11 were not run (GPT-OSS-20B on the five benchmarks other than
HMMT, and MBPP for the six models other than GPT-OSS-20B and the two 8B
models). The primary roster therefore contains 38 nondegenerate
main-matrix cells, one GPT-OSS HMMT cell, and two MBPP cells. The two Putnam rows have $n=23$ and are displayed
as exploratory, producing 43 displayed rows but 41 primary rows. Aggregates
weight cells equally; weighting each primary cell by its number of problems
gives $\EqThreeGapNWeightedPP\pp$ instead of $\EqThreeGapPP\pp$. Both Gini
coefficients use the standard sorted-vector formula on utilities clipped at
zero over eligible positions. The pooled Gini includes every chain of a cell;
the within-chain mean omits chains whose clipped eligible utility is zero
everywhere, for which the Gini is undefined.

Thirteen primary cells were collected again with per-request sampling seeds,
and two 8B MATH-500 rows use their full $n=500$ membership.

\MeasurementTable

\begin{longtable}{llrrrr}
\caption{\textbf{Per-cell measurement record.} Putnam rows are exploratory and are excluded from the 41-cell primary aggregate. The max-utility column averages each example's largest observed utility, a different estimand from Eq.~(2) that can fall below it because utility subtracts the base outcome.}\label{tab:all-cells}\\
\toprule
Model & Benchmark & $n$ & Eq.~(2) pp & pooled Gini & max-utility pp \\
\midrule
\endfirsthead
\caption[]{(continued)}\\
\toprule
Model & Benchmark & $n$ & Eq.~(2) pp & pooled Gini & max-utility pp \\
\midrule
\endhead
DS-R1-Llama-8B & MATH-500 & 500 & +14.69 & 0.778 & +28.90 \\
DS-R1-Llama-8B & GPQA & 198 & +17.20 & 0.836 & +25.63 \\
DS-R1-Llama-8B & HumanEval & 164 & +9.87 & 0.854 & +19.82 \\
DS-R1-Llama-8B & AIME 2025 & 30 & +2.77 & 0.984 & +2.50 \\
DS-R1-Llama-8B & HMMT & 30 & +9.83 & 0.928 & +11.67 \\
Nemotron-Nano-8B & MATH-500 & 500 & +16.17 & 0.760 & +29.60 \\
Nemotron-Nano-8B & HumanEval & 164 & +13.91 & 0.799 & +33.08 \\
Nemotron-Nano-8B & AIME 2025 & 30 & +1.82 & 0.941 & +3.33 \\
Nemotron-Nano-8B & HMMT & 30 & +8.39 & 0.852 & +13.33 \\
DS-R1-Qwen-14B & MATH-500 & 500 & +19.37 & 0.775 & +39.55 \\
DS-R1-Qwen-14B & GPQA & 198 & +26.81 & 0.845 & +39.14 \\
DS-R1-Qwen-14B & HumanEval & 164 & +19.42 & 0.803 & +38.72 \\
DS-R1-Qwen-14B & AIME 2025 & 30 & +8.71 & 0.939 & +16.67 \\
DS-R1-Qwen-14B & HMMT & 30 & +21.35 & 0.911 & +31.67 \\
GLM-4.5-Air & MATH-500 & 500 & +33.01 & 0.819 & +48.00 \\
GLM-4.5-Air & GPQA & 198 & +30.40 & 0.859 & +46.46 \\
GLM-4.5-Air & HumanEval & 164 & +20.01 & 0.826 & +35.82 \\
GLM-4.5-Air & AIME 2025 & 30 & +5.00 & 0.971 & +11.67 \\
GLM-4.5-Air & HMMT & 30 & +19.20 & 0.948 & +25.00 \\
Gemma-3-27B & MATH-500 & 500 & +6.17 & 0.953 & +8.50 \\
Gemma-3-27B & GPQA & 198 & +23.94 & 0.893 & +26.01 \\
Gemma-3-27B & HumanEval & 164 & +16.16 & 0.874 & +20.27 \\
Gemma-3-27B & AIME 2025 & 30 & +7.14 & 0.970 & +10.00 \\
Gemma-3-27B & HMMT & 30 & +18.66 & 0.929 & +27.50 \\
Qwen3-32B & MATH-500 & 500 & +29.27 & 0.815 & +42.25 \\
Qwen3-32B & GPQA & 168 & +32.14 & 0.887 & +34.38 \\
Qwen3-32B & HumanEval & 60 & +16.67 & 0.867 & +26.67 \\
Qwen3-32B & AIME 2025 & 30 & +3.65 & 0.950 & +7.50 \\
Qwen3-32B & HMMT & 30 & +13.87 & 0.916 & +23.33 \\
Llama-Nemotron-70B & MATH-500 & 500 & +15.62 & 0.931 & +17.50 \\
Llama-Nemotron-70B & GPQA & 198 & +33.11 & 0.856 & +36.36 \\
Llama-Nemotron-70B & HumanEval & 164 & +7.07 & 0.958 & +6.71 \\
Llama-Nemotron-70B & AIME 2025 & 30 & +4.21 & 0.981 & +6.67 \\
Mistral-Large-2411 & MATH-500 & 500 & +13.94 & 0.934 & +15.95 \\
Mistral-Large-2411 & GPQA & 198 & +18.33 & 0.906 & +13.64 \\
Mistral-Large-2411 & HumanEval & 164 & +19.32 & 0.984 & +3.05 \\
Mistral-Large-2411 & AIME 2025 & 30 & +8.87 & 0.937 & +11.67 \\
Mistral-Large-2411 & HMMT & 30 & +7.29 & 0.967 & +13.33 \\
GPT-OSS-20B & HMMT & 30 & +7.49 & 0.876 & +16.67 \\
DS-R1-Llama-8B & MBPP & 105 & +2.32 & 0.979 & +2.38 \\
Nemotron-Nano-8B & MBPP & 106 & +3.44 & 0.962 & +4.25 \\
DS-R1-Llama-8B & Putnam$^{\dagger}$ & 23 & +11.53 & 0.843 & +20.65 \\
Nemotron-Nano-8B & Putnam$^{\dagger}$ & 23 & +14.36 & 0.798 & +29.35 \\
\bottomrule
\multicolumn{6}{l}{\footnotesize $^{\dagger}$ Exploratory $n=23$ row, displayed but excluded from primary aggregates.}\\
\end{longtable}

\FloatBarrier
\section{Held-Out Scoring Audits}
\label{sec:independent-audit}

\IndependentScoringTable

The fresh-eight same-draw row uses the same eight new outcomes to select and
score and therefore remains selected on noise. The stored-winner row freezes
the original four-draw choice and scores it on eight unseen continuations. The
fresh 4/4 row selects on four fresh continuations and scores on the
complementary four. The 2/2 matrix audit exhausts all three complementary
partitions and both directions, with analytic tie-neutral inclusion at the
budget cutoff. Each protocol estimates a different selector, so we report them separately.

Pooled Gini itself is reproducible across paired halves, with Pearson range
from $\CrossfitGiniReliabilityMin$ to $\CrossfitGiniReliabilityMax$. The median
disjoint-half Gini-gap correlation is $\CrossfitMedianGiniGap$
$[\CrossfitMedianGiniGapLow,\CrossfitMedianGiniGapHigh]$.

\MeasurementRobustnessTable

The separately generated 4/4 audit (Table~\ref{tab:measurement-robustness})
covers $\IndependentPrefixExamples$
examples in $\IndependentPrefixCells$ cells and gives
$\IndependentPrefixGapPP\pp$
$[\IndependentPrefixGapLowPP,\IndependentPrefixGapHighPP]$, positive in every
cell. Its cross-sample cell-Gini reliability is
$r=\IndependentPrefixGiniReliability$. The boundary-density controls change the
candidate set and thereby the magnitude of same-draw opportunity; they do not
test semantic segmentation.

\FloatBarrier
\section{Policy-Class Protocol}
\label{sec:policy-protocol}

The declared depth class contains 21 policies indexed by a target relative
depth $q\in\{0,0.05,\ldots,1\}$: policy $q$ selects the eligible positions whose
relative depth is closest to $q$ (ties to the earlier position), and $q$ is
chosen by five-fold cross-validation on the selection sample. Candidate
selection uses one continuation sample and final scoring uses the other; the
direction is then reversed. Sample A is the point restart at position $j$ and
sample B is the radius-one block restart centered at $j+1$, which regenerates
from the same prefix; both are separately generated with identical sampling
settings and graders. Candidates are positions $1$ to $K_i-3$, where both
samples exist. The
enriched sensitivity analysis nests selection over the original grid and
regularized length-by-rank tables. In both cases, continuation outcomes remain in the
oracle selection rule. These experiments therefore quantify residual
opportunity, not predictability from features.

\DepthResidualTable
\RichResidualTable

The evaluated router uses within-example engineered text and structure
features together with PRM and token-log-probability summaries, a listwise
objective, a learned depth prior, positional support, and a confidence gate. Attribution rests on the paired always-last
controls.

The evaluated router is a two-hidden-layer MLP with widths 64 and 32, GELU
activations, and dropout 0.10 after the first hidden layer. It is fitted with
AdamW at learning rate $3\times10^{-4}$,
weight decay $10^{-4}$, mini-batch size 256, and at most 80 epochs. Early
stopping uses patience 12. Each split keeps all positions from one example on
the same side, reserves 20\% of examples for outer evaluation, and reserves
15\% of the remaining training examples for inner validation. The evaluated
configuration uses ListNet temperature 1, a three-bin depth-prior floor of
0.1, and four retained outcomes per action. The multi-seed controls repeat the
complete split-and-fit procedure at their reported seeds; no test outcome is
used to choose a threshold or architecture.
The confidence gate is a small MLP over chain-level structure, text, and PRM
summary features, trained to predict base-chain correctness. Its threshold is
selected on inner validation, maximizing mean GR across budgets subject to
routing at least 10\% of chains, and is then frozen for testing. On held-out problems the gate routes a mean of
$\GateRoutedMeanPct\%$ of chains, ranging from $\GateRoutedMinPct\%$ to
$\GateRoutedMaxPct\%$ across split seeds, so its operating point varies
substantially between splits.

\DecisionTable
\RequestedComponentTable

The leave-domain-out control (Table~\ref{tab:domain-transfer}) removes all
target-domain benchmarks from fitting
and validation, disables benchmark identity, and applies a shared late-only
support. Its positive contrasts are against unconstrained uniform placement;
paired differences from always-last remain unresolved. The control therefore
establishes within-suite transfer of the late-support policy, not general
content-based restart localization.

\DomainTransferTable

\FloatBarrier
\section{Exact Self-Consistency Frontier Protocol}
\label{sec:frontier-protocol}

The panel is the first 100 MATH-500 problems with identifiers sorted as
strings, run with
three generation seeds and an 8192-token cap per request. SC returns the
plurality parsed answer, with ties broken toward the earliest sample;
unparsable samples do not vote but their tokens count, and a pool with no
parsable answer is scored incorrect. Generated output counts every generated
token, including the fresh base chain of always-last and the router. For each
problem and seed, all subsets for SC(1) to SC(4) are enumerated. For a target aggregate output budget, a linear
program chooses the globally randomized mixture over adjacent frontier points.
Policy accuracy is paired with the same problem's frontier construction, and
the percentile bootstrap resamples problem clusters with their seeds intact.
The four integer SC points are
$(\SCKOneRatio,\SCKOneAccuracy)$,
$(\SCKTwoRatio,\SCKTwoAccuracy)$,
$(\SCKThreeRatio,\SCKThreeAccuracy)$, and
$(\SCKFourRatio,\SCKFourAccuracy)$. These results are conditional on the frozen
DeepSeek-R1-Distill-Qwen-14B/MATH-500 panel.

\FrontierTable

The two-model MATH result is a fixed-panel sensitivity using the two available
cells from a planned four-cell model-by-benchmark panel. The Qwen3-32B cell alone is
unresolved, and the missing GPQA cells prevent the planned panel from being
identified. The PRM best-of-$N$, tree, and H200 rows of
Table~\ref{tab:efficiency-controls} are reported with their
own cohorts and ledgers rather than pooled with the exact DeepSeek
frontier result.

\EfficiencyControlTable

\FloatBarrier
\section{Loss and Feature Diagnostics}
\label{sec:loss-appendix}

The primary fair-label comparison freezes 24 split seeds and bootstraps
problem clusters. The unmasked problem-population interval crosses zero even
though a separate seed-stability interval is positive; the main text uses the
problem-cluster result for inference. Inside the benchmark-conditional late
support, both
problem- and seed-level contrasts are unresolved. We do not use an equivalence
claim for this primary comparison because its protocol did not freeze and power
an equivalence margin.

\LossExtensionTable

The extension froze the additional seed labels, code, data, and configuration
before running seeds 24 to 63, but after the original 24-seed artifact was
available. Under its frozen decision rule, the mask-free contrast is positive under both
its problem-cluster interval (Table~\ref{tab:loss-extension}) and its paired-seed
interval $[\FairLossExtendedUnmaskedSeedLow,\FairLossExtendedUnmaskedSeedHigh]$,
and the contrast inside the benchmark-conditional late support is equivalent
within the frozen $\pm\FairLossEquivalenceMargin$ margin (paired-seed interval
$[\FairLossExtendedInBandSeedLow,\FairLossExtendedInBandSeedHigh]$). This is post-lock
stability evidence rather than independent confirmation; the primary inference
remains the 24-seed problem-cluster interval.

Direct feature additions under the evaluated objective show no detected
improvement: Math-Shepherd
PRM summaries~\citep{math_shepherd}, token log-probabilities, and hidden-state representations do not
improve the evaluated selector on matched coverage. PRM best-of-$N$ and a
shallow PRM tree also do not beat self-consistency in their scoped controls.
These diagnostics identify the current bottleneck without excluding other
representations, objectives, or search algorithms.

\FloatBarrier
\section{Additional Analyses and Scope}
\label{sec:nonpromoted}

A predeclared feasible-spend analysis did not pass its randomization gate, so
it establishes no learned-spend effect. Provider total-token accounting and
additional always-last analyses are reported as diagnostics. The
domain-transfer, H200 timing, tree-search, and fixed-panel controls are
reported separately because their cohorts, cost ledgers, or uncertainty units
differ from the primary results.

\FloatBarrier
\section{Reproducibility and Release}
\label{sec:reproducibility}

The per-cell measurement records, utility maps, 41-cell inclusion record,
router split seeds, and the JSON certificates used by this manuscript are
available from the authors on request, together with a hash registry and a
script that verifies every registered input and regenerates the tables.
Figures are compiled directly from TikZ/PGFPlots.

For the 28 historical primary rows, settings are reconstructed from preserved
commands and logs; the original runs did not embed per-cell manifests, so these
settings are not cryptographically bound. They are temperature $0.7$,
top-$p=0.95$, no explicit top-$k$ restriction, and a $1024$-token generation
cap ($1280$ for Qwen3-32B/HumanEval and $4096$ for the legacy Mistral-Large
GPQA and HumanEval rows). The 13 recollected rows
use temperature $0.7$, top-$p=0.95$, no explicit top-$k$ restriction, a
$1024$-token cap, and four distinct deterministic request seeds. Fresh
exact-prefix audits reuse the frozen original base chains and prefixes rather
than regenerating them; only their new continuations use the $1024$-token audit
protocol. All primary measurement statistics use four retained continuation
outcomes per action, except Qwen3-32B/HumanEval (two) and the legacy
Mistral-Large GPQA and HumanEval rows (eight). The separate exact-frontier assay uses a newly generated,
$8192$-token-cap protocol. Repository checkpoint identifiers, benchmark
versions, prompts, local graders, candidate segmentation, seeds, and retry
rules are recorded for the recollected and later collections.
MATH-500, GPQA Diamond,
HumanEval, AIME 2025, HMMT, MBPP, and Putnam use their published problem
sets or splits (per-cell problem counts in Table~\ref{tab:all-cells}) and
pinned local graders. Numeric tasks use normalized extracted
answers, GPQA uses parsed multiple-choice indices, and code tasks execute the
published tests against extracted Python code. Commands for the
measurement aggregation, held-out scoring
audits, policy decomposition, positional residuals, tie-resolution analysis,
and exact SC frontier are also available on request.

\paragraph{Checkpoints and execution}
\begingroup\raggedright
The evaluated repository identifiers are
\nolinkurl{deepseek-ai/DeepSeek-R1-Distill-Llama-8B}~\citep{deepseek_r1_2025},
\nolinkurl{nvidia/Llama-3.1-Nemotron-Nano-8B-v1}~\citep{nvidia_nemotron_nano_8b_2025},
\nolinkurl{deepseek-ai/DeepSeek-R1-Distill-Qwen-14B}~\citep{deepseek_r1_2025}, \nolinkurl{openai/gpt-oss-20b}~\citep{gpt_oss_20b_2025}, \nolinkurl{zai-org/GLM-4.5-Air}~\citep{glm_4_5_air_2025}, \nolinkurl{google/gemma-3-27b-it}~\citep{gemma3_2025},
\nolinkurl{Qwen/Qwen3-32B}~\citep{qwen3_2025},
\nolinkurl{nvidia/Llama-3.1-Nemotron-70B-Instruct-HF}~\citep{nvidia_nemotron_70b_2024}, and
\nolinkurl{mistralai/Mistral-Large-Instruct-2411}~\citep{mistral_large_2411}. Repository revisions are reported where preserved; the historical collections
do not record a revision for every checkpoint.
The historical and recollected campaigns use model-compatible multi-GPU
inference in bfloat16 except for the native quantized GPT-OSS checkpoint. Exact per-model tensor-parallel assignments are not recorded for every run.
The recollected and later campaigns retain
campaign-specific seeds, generation settings, token counts, and available job
metadata. Older rows lack uniform per-action manifests, so row-level provenance for the
historical sweep is incomplete. Router fitting is CPU-scale;
continuation collection is the GPU-dominant stage. The exact-frontier
experiment uses three independently generated seeds, and the order-balanced
timing assay uses the same H200 hardware for both arms.
\par\endgroup

\paragraph{Asset and assistance disclosure}
The paper identifies the evaluated model repositories and benchmark datasets.
We do not provide a complete per-asset license inventory, as reflected in
the checklist. No human-subject data are collected.
Language models were used for prose editing and code review; the authors
verified every generated table, figure, and numerical statement against the
frozen sources and hash manifest.

\end{document}